\def\year{2019}\relax
\documentclass[letterpaper]{article} 
\usepackage{aaai19}  
\usepackage{times}  
\usepackage{helvet}  
\usepackage{courier}  
\usepackage{url}  
\usepackage{graphicx}  
\usepackage{subfig}
\usepackage{bm}
\usepackage{tabularx}                             
\usepackage{algorithm}
\usepackage{algorithmicx}
\usepackage{amssymb}
\usepackage[noend]{algpseudocode}
\usepackage{multicol}
\usepackage{float}
\usepackage{amsmath}
\usepackage{listings}
\usepackage{pdfpages}
\newcolumntype{L}[1]{>{\raggedright\arraybackslash}m{#1}}
\newcolumntype{P}[1]{>{\centering\arraybackslash}p{#1}}

\begin{document}
%
\title{Memory Bounded Open-Loop Planning in Large POMDPs using\\ Thompson Sampling}
\author{Thomy Phan\\
LMU Munich\\
thomy.phan@ifi.lmu.de
\And
Lenz Belzner\\
MaibornWolff\\
lenz.belzner@maibornwolff.de
\And
Marie Kiermeier\\
LMU Munich\\
marie.kiermeier@ifi.lmu.de
\AND
Markus Friedrich\\
LMU Munich\\
markus.friedrich@ifi.lmu.de
\And
Kyrill Schmid\\
LMU Munich\\
kyrill.schmid@ifi.lmu.de
\And
Claudia Linnhoff-Popien\\
LMU Munich\\
linnhoff@ifi.lmu.de
}

\maketitle
\begin{abstract}
State-of-the-art approaches to partially observable planning like POMCP are based on stochastic tree search. While these approaches are computationally efficient, they may still construct search trees of considerable size, which could limit the performance due to restricted memory resources. In this paper, we propose \emph{Partially Observable Stacked Thompson Sampling (POSTS)}, a memory bounded approach to open-loop planning in large POMDPs, which optimizes a fixed size stack of Thompson Sampling bandits. We empirically evaluate POSTS in four large benchmark problems and compare its performance with different tree-based approaches. We show that POSTS achieves competitive performance compared to tree-based open-loop planning and offers a performance-memory tradeoff, making it suitable for partially observable planning with highly restricted computational and memory resources.
\end{abstract}

\section{Introduction}

Many real-world problems can be modeled as \emph{Partially Observable Markov Decision Process (POMDP)}, where the true state is unknown to the agent due to limited and noisy sensors. The agent has to reason about the \emph{history} of past observations and actions, and maintain a \emph{belief state} as a distribution of possible states. POMDPs have been widely used to model decision making problems in the context of planning and reinforcement learning \cite{ross2008online}.

Solving POMDPs exactly is computationally intractable for domains with enormous state spaces and long planning horizons. First, the space of possible belief states grows exponentially w.r.t. the number of states $N$, since that space is $N$-dimensional, which is known as \emph{curse of dimensionality} \cite{kaelbling1998planning}. Second, the number of possible histories grows exponentially w.r.t. the horizon length, which is known as the \emph{curse of history} \cite{pineau2006anytime}.

In the last few years, \emph{Monte-Carlo planning} has been proposed to break both curses with statistical sampling. These methods construct sparse trees over belief states and actions, representing the state-of-the-art for efficient planning in large POMDPs \cite{silver2010monte,somani2013despot,bai2014thompson}. While these approaches avoid exhaustive search, the constructed \emph{closed-loop} trees can still become arbitrarily large for highly complex domains, which could limit the performance due to restricted memory resources \cite{powley2017memory}. In contrast, \emph{open-loop} approaches only focus on searching action sequences and are independent of the history and belief state space. Open-loop approaches are able to achieve competitive performance compared to closed-loop planning, when the problem is too large to provide sufficient computational and memory resources \cite{weinstein2013open,perez2015open,lecarpentier2018open}. However, open-loop planning has been a less popular choice for decision making in POMDPs so far \cite{yu2005open}.

In this paper, we propose \emph{Partially Observable Stacked Thompson Sampling (POSTS)}, a memory bounded approach to open-loop planning in large POMDPs, which optimizes a fixed size stack of Thompson Sampling bandits.

To evaluate the effectiveness of POSTS, we formulate a tree-based approach, called \emph{Partially Observable Open-Loop Thompson Sampling (POOLTS)} and show that POOLTS is able to find optimal open-loop plans with sufficient computational and memory resources.

We empirically test POSTS in four large benchmark problems and compare its performance with POOLTS and other tree-based approaches like POMCP. We show that POSTS achieves competitive performance compared to tree-based open-loop planning and offers a performance-memory tradeoff, making it suitable for partially observable planning with highly restricted computational and memory resources.

\section{Background}

\subsection{Partially Observable Markov Decision Processes}

A POMDP is defined by a tuple $M = \langle\mathcal{S},\mathcal{A},\mathcal{P},\mathcal{R},\mathcal{O},\Omega, b_{0} \rangle$, where $\mathcal{S}$ is a (finite) set of states, $\mathcal{A}$ is the (finite) set of actions, $\mathcal{P}(s_{t+1}|s_{t}, a_{t})$ is the transition probability function, $\mathcal{R}(s_{t}, a_{t})$ is the scalar reward function, $\mathcal{O}$ is a (finite) set of observations, $\Omega(o_{t+1}|s_{t+1}, a_{t})$ is the observation probability function, and $b_{0}$ is a probability distribution over initial states $s_{0} \in \mathcal{S}$. It is always assumed, that $s_{t}, s_{t+1} \in \mathcal{S}$, $a_{t} \in \mathcal{A}$, and $o_{t}, o_{t+1} \in \mathcal{O}$ at time step $t$.

A \emph{history} $h_{t} = \big[a_{0},o_{1},...,a_{t-1},o_{t}\big]$ is a sequence of actions and observations. A \emph{belief state} $b_{h_{t}}(s_{t})$ is a sufficient statistic for history $h_{t}$ and defines a probability distribution over states $s_{t}$ given $h_{t}$. $\mathcal{B}$ is the space of all possible belief states. $b_{0}$ represents the \emph{initial belief state} $b_{h_{0}}$. The belief state can be updated by Bayes theorem:
\begin{equation}\label{eq:belief_update}
b_{h_{t}}(s_{t}) = \eta \Omega(o_{t}|s_{t}, a) \sum_{s \in \mathcal{S}}^{} \mathcal{P}(s_{t}|s, a) b_{h}(s_{t})
\end{equation}
where $\eta = \frac{1}{\Omega(o_{t+1}|b_{h}, a)}$ is a normalizing constant, $a = a_{t-1}$ is the last action, and $h$ is the history without $a$ and $o_{t}$.

The goal is to find a \emph{policy} $\pi : \mathcal{B} \rightarrow \mathcal{A}$, which maximizes the return $G_{t}$ at state $s_{t}$ for a horizon $T$:
\begin{equation}\label{eq:return}
G_{t} = \sum_{k=0}^{T-1} \gamma^{k} \cdot \mathcal{R}(s_{t+k}, a_{t+k})
\end{equation}
where $\gamma \in [0,1]$ is the discount factor. If $\gamma < 1$, then present rewards are weighted more than future rewards.

The \emph{value function} $V^{\pi}(b_{t}) = \mathbb{E}_{\pi}[G_{t}|b_{t}\big]$ is the expected return conditioned on belief states given a policy $\pi$. An optimal policy $\pi^{*}$ has a value function $V^{\pi^{*}}= V^{*}$ with $V^{*}(b_{t}) \geq V^{\pi'}(b_{t})$ for all $b_{t} \in \mathcal{B}$ and $\pi' \neq \pi^{*}$.

\subsection{Multi-armed Bandits}
\emph{Multi-armed Bandits (MABs or bandits)} are fundamental decision making problems, where an agent has to repeatedly select an arm among a given set of arms in order to maximize its future pay-off. MABs can be considered as problems with a single state $s$, a set of actions $a \in \mathcal{A}$, and a stochastic reward function $\mathcal{R}(s, a) := X_{a}$, where $X_{a}$ is a random variable with an unknown distribution $f_{X_{a}}(x)$. To solve a MAB, one has to determine the action, which maximizes the expected reward $\mathbb{E}\big[X_{a}\big]$. The agent has to balance between sufficiently trying out actions to accurately estimate their expected reward and to exploit its current knowledge on all arms by selecting the arm with the currently highest expected reward. This is known as the \emph{exploration-exploitation dilemma}, where exploration can lead to actions with possibly higher rewards but requires time for trying them out, while exploitation can lead to fast convergence but possibly gets stuck in a local optimum. In this paper, we will cover UCB1 and Thompson Sampling as MAB algorithms.

\subsubsection{UCB1}
In \emph{UCB1}, actions are selected by maximizing the upper confidence bound of action values $\textit{UCB1}(a) = \overline{X_{a}} + c \sqrt{\frac{\textit{log}(N_{\textit{total}})}{N_{a}}}$, where $\overline{X_{a}}$ is the current average reward when choosing $a$, $c$ is an exploration constant, $N_{\textit{total}}$ is the total number of action selections, and $N_{a}$ is the number of times action $a$ was selected. The second term represents the exploration bonus, which becomes smaller with increasing $N_{a}$ \cite{auer2002finite}.

UCB1 is a popular MAB algorithm and widely used in various challenging domains \cite{kocsis2006bandit,bubeck2010open,silver2016mastering,silver2017mastering}.

\subsubsection{Thompson Sampling}
\emph{Thompson Sampling} is a Bayesian approach to balance between exploration and exploitation of actions \cite{thompson1933likelihood}. The unknown reward distribution of $X_{a}$ of each action $a \in \mathcal{A}$ is modeled by a parametrized likelihood function $P_{a}(x|\theta)$ with a parameter vector $\theta$. Given a prior distribution $P_{a}(\theta)$ and a set of past observed rewards $D_{a} = \{x_{a,1},x_{a,2},...,x_{a,N_{a}}\}$, the posterior distribution $P_{a}(\theta|D_{a})$ can be inferred by using Bayes rule $P_{a}(\theta|D_{a}) \propto \prod_{i}^{}P_{a}(x_{a,i}|\theta)P_{a}(\theta)$. The expected reward of each action $a \in \mathcal{A}$ can be estimated by sampling $\theta \sim P_{a}(\theta|D_{a})$ from the posterior. The action with the highest sampled expected reward $\mathbb{E}_{\theta}\big[X_{a}\big]$ is selected.

Thompson Sampling has been shown to be an effective and robust algorithm for making decisions under uncertainty \cite{chapelle2011empirical,kaufmann2012thompson,bai2013bayesian,bai2014thompson}.

\subsection{Planning in POMDPs}
\emph{Planning} searches for an (near-)optimal policy given a model $\hat{M}$ of the environment $M$, which usually consists of explicit probability distributions of the POMDP. Unlike \emph{offline planning}, which searches the whole (belief) state space to find the optimal policy $\pi^{*}$, \emph{local planning} only focuses on finding a policy $\pi_{t}$ for the current (belief) state by taking possible future (belief) states into account \cite{weinstein2013open}. Thus, local planning can be applied \emph{online} at every time step at the current state to recommend the next action for execution. Local planning is usually restricted to a time or computation budget $n_{\textit{b}}$ due to strict real-time constraints \cite{bubeck2010open,weinstein2013open,perez2015open}.

In this paper, we focus on local \emph{Monte-Carlo planning}, where $\hat{M}$ is a generative model, which can be used as black box simulator \cite{kocsis2006bandit,silver2010monte,weinstein2013open,bai2014thompson}. Given $s_{t}$ and $a_{t}$, the simulator $\hat{M}$ provides a sample $\langle s_{t+1},o_{t+1},r_{t}\rangle \sim \hat{M}(s_{t},a_{t})$. Monte-Carlo planning algorithms can approximate $\pi^{*}$ and $V^{*}$ by iteratively simulating and evaluating action sequences without reasoning about explicit probability distributions of the POMDP.

\begin{figure}[!ht]
     \subfloat[closed-loop tree\label{fig:closed_loop_planning}]{%
       \includegraphics[width=0.2\textwidth]{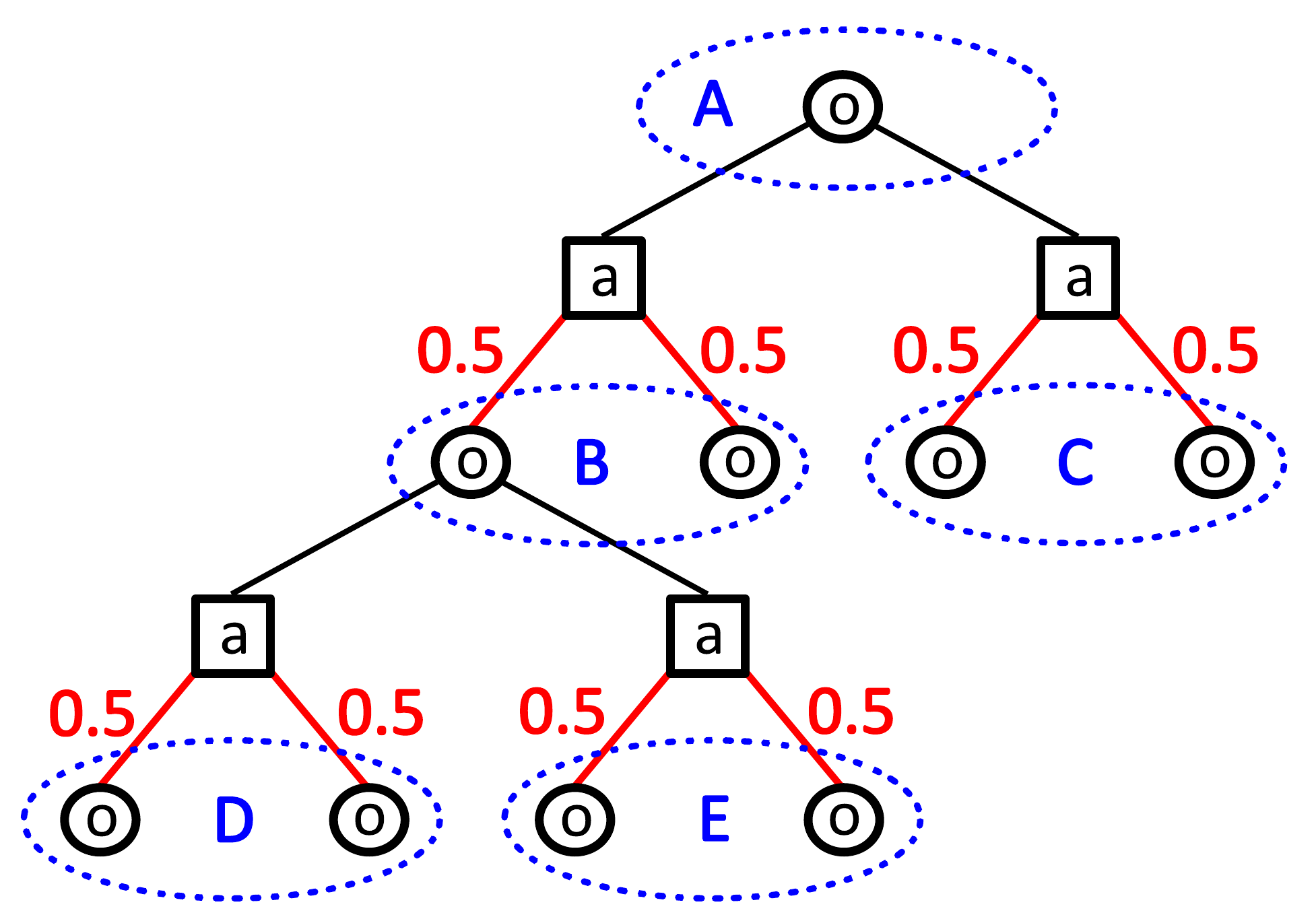}
     }
     \hfill
     \subfloat[open-loop tree\label{fig:open_loop_planning}]{%
       \includegraphics[width=0.15\textwidth]{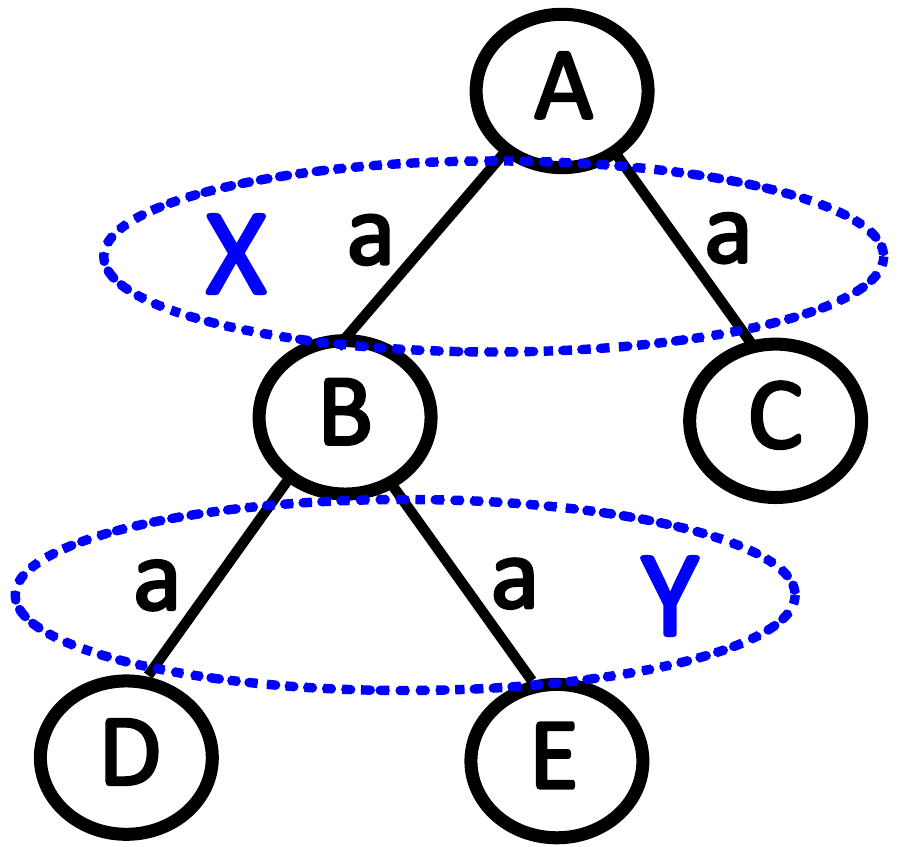}
     }\hfill
     \subfloat[stacked\label{fig:open_loop_planning_compressed}]{%
       \includegraphics[width=0.1\textwidth]{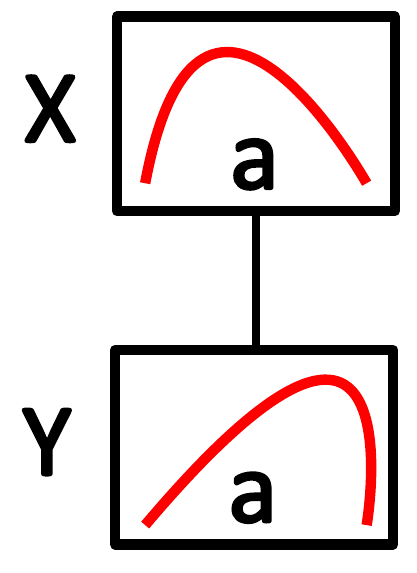}
     }
     \caption{Illustration of closed- and open-loop planning schemes. (a) Closed-loop tree with state observations (circular nodes) and actions (rectangular nodes). Red links correspond to stochastic observations made with a probability of 0.5. (b) Open-loop tree with links as actions and history distribution nodes according to the blue dotted ellipses in Fig. \ref{fig:closed_loop_planning}. (c) Open-loop approach with stack of action distributions according to the blue dotted ellipses in Fig. \ref{fig:open_loop_planning}.}
     \label{fig:closed_vs_open_loop_planning}
\end{figure}

Local planning can be closed- or open-loop. \emph{Closed-loop planning} conditions the action selection on histories of actions and observations. \emph{Open-loop planning} only conditions the action selection on previous sequences of actions $p_{T} = [a_{1},...,a_{T}]$ (also called \emph{open-loop plans} or simply \emph{plans}) and summarized statistics about predecessor (belief) states \cite{bubeck2010open,weinstein2013open,perez2015open}. An example is shown in Fig. \ref{fig:closed_vs_open_loop_planning}. A closed-loop tree for a domain with $\Omega(s_{t+1}|s_{t},a_{t}) = 0.5$ is shown in Fig. \ref{fig:closed_loop_planning}, while Fig. \ref{fig:open_loop_planning} shows the corresponding open-loop tree which summarizes the observation nodes of Fig. \ref{fig:closed_loop_planning} within the blue dotted ellipses into history distribution nodes. Open-loop planning can be further simplified by only regarding statistics about the expected return of actions at specific time steps (Fig. \ref{fig:open_loop_planning_compressed}). In that case, only a stack of $T$ statistics is used to sample plans for simulation and evaluation \cite{weinstein2013open}.

\emph{Partially Observable Monte-Carlo Planning (POMCP)} is a closed-loop approach based on \emph{Monte-Carlo Tree Search (MCTS)} \cite{silver2010monte}. POMCP uses a search tree of histories with \textit{o-nodes} representing observations and \textit{a-nodes} representing actions (Fig. \ref{fig:closed_loop_planning}). Each \textit{o-node} has a visit count $N(h_{t})$ and a value estimate $V(h_{t}) = V(b_{t})$ for history $h_{t}$ and belief state $b_{t}$. Each \textit{a-node} has a visit count $N(h_{t})$ and a value estimate $Q(h_{t},a_{t}) = \mathbb{E}[G_{t}|h_{t},a_{t}]$ for action $a_{t}$ and history $h_{t}$. A simulation starts at the current belief state and is divided into two stages: In the first stage, a tree policy $\pi_{\textit{tree}}$ is used to traverse the tree until a leaf node is reached. Actions are selected via $\pi_{\textit{tree}}(h_{t})$ and simulated in $\hat{M}$ to determine the next nodes to visit. $\pi_{\textit{tree}}$ can be implemented with MABs, where each \textit{o-node} represents a MAB. In the second stage, a rollout policy $\pi_{\textit{rollout}}$ is used to sample action sequences until a terminal state or a maximum search depth $T$ is reached. The observed rewards are accumulated to returns $G_{t}$ (Eq. \ref{eq:return}), propagated back to update the value estimate of every node in the simulated path, and a new leaf node is added to the search tree. $\pi_{\textit{rollout}}$ can be used to integrate domain knowledge into the planning process to focus the search on promising states \cite{silver2010monte}. The original version of POMCP uses UCB1 for $\pi_{\textit{tree}}$ and is shown to converge to the optimal best-first tree with sufficient computation \cite{silver2010monte}.

\cite{lecarpentier2018open} formulates an open-loop variant of MCTS using UCB1 as $\pi_{\textit{tree}}$, called \emph{Open-Loop Upper Confidence bound for Trees (OLUCT)}, which could be easily extended to POMDPs by constructing a tree, which summarizes all \textit{o-nodes} to history distribution nodes (Fig. \ref{fig:open_loop_planning}).

Open-loop planning generally converges to sub-optimal solutions in stochastic domains, since it ignores (belief) state values $V(b_{t})$ and optimizes the node values $V(N_{t})$ (Fig. \ref{fig:open_loop_planning}) instead \cite{lecarpentier2018open}. If the problem is too complex to provide sufficient computation budget $n_{b}$ or memory capacity, then open-loop approaches are competitive to closed-loop approaches, since they need to explore a much smaller search space to converge to an appropriate solution \cite{weinstein2013open,lecarpentier2018open}.

\section{Related Work}
Tree-based approaches to open-loop planning condition the action selection on previous action sequences as shown in Fig. \ref{fig:open_loop_planning} \cite{bubeck2010open,perez2015open,lecarpentier2018open}. Such approaches have been thoroughly evaluated for fully observable problems, but have been less popular for partially observable problems so far \cite{yu2005open}. POSTS is based on stacked open-loop planning, where a stack of $T$ distributions over actions is maintained to generate open-loop plans with high expected return \cite{weinstein2013open,belzner2017stacked}. Unlike previous approaches, POSTS is a \emph{memory-bounded} open-loop approach to \emph{partially observable planning}.

\cite{yu2005open} proposed an open-loop approach to planning in POMDPs by using hierarchical planning. An open-loop plan is constructed at an abstract level, where uncertainty w.r.t. particular actions is ignored. A low-level planner controls the actual execution by explicitly dealing with uncertainty. POSTS is more general, since it performs planning directly on the \emph{original problem} and does not require the POMDP to be transformed for hierarchical planning.

\cite{powley2017memory} proposed a memory bounded version of MCTS with a state pool to add, discard, or reuse states depending on their visitation frequency. However, this approach cannot be easily adapted to tree-based open-loop approaches, because it requires (belief) states to be identifiable. POSTS does not require a pool to reuse states or nodes, but only maintains a \emph{fixed size stack} of Thompson Sampling bandits, which adapt according to the \emph{temporal dependencies} between actions.


\section{Open-Loop Search with Thompson Sampling}
\subsubsection{Generalized Thompson Sampling}
We use a variant of Thompson Sampling, which works for arbitrary reward distributions as proposed in \cite{bai2013bayesian,bai2014thompson} by assuming that $X_{a_{t}}$ follows a Normal distribution $\mathcal{N}(\mu, \frac{1}{\tau})$ with unknown mean $\mu$ and precision $\tau = \frac{1}{\sigma^2}$, where $\sigma^2$ is the variance. $\langle\mu, \tau\rangle$ follows a Normal Gamma distribution $\mathcal{NG}(\mu_{0},\lambda,\alpha,\beta)$ with $\lambda > 0$, $\alpha \geq 1$, and $\beta \geq 0$. The distribution over $\tau$ is a Gamma distribution $\tau \sim \textit{Gamma}(\alpha,\beta)$ and the conditional distribution over $\mu$ given $\tau$ is a Normal distribution $\mu \sim \mathcal{N}(\mu_{0},\frac{1}{\lambda\tau})$.

Given a prior distribution $P(\theta) = \mathcal{NG}(\mu_{0},\lambda_{0},\alpha_{0},\beta_{0})$ and $n$ observations $D = \{x_{1},...,x_{n}\}$, the posterior distribution is defined by $P(\theta|D) = \mathcal{NG}(\mu_{1},\lambda_{1},\alpha_{1},\beta_{1})$, where $\mu_{1} = \frac{\lambda_{0}\mu_{0} + n\overline{X}}{\lambda_{0}+n}$,
$\lambda_{1} = \lambda_{0}+n$, $\alpha_{1} = \alpha_{0}+\frac{n}{2}$, and $\beta_{1} = \beta_{0}+\frac{1}{2}(n \sigma^{2} + \frac{\lambda_{0}n(\overline{X}-\mu_{0})^2}{\lambda_{0}+n})$. $\overline{X}$ is the mean of all values in $D$ and $\sigma^{2} = \frac{1}{n}\sum_{i = 1}^{n} (x_{i} - \overline{X})^2$ is the variance.

The posterior is inferred for each action $a_{t} \in \mathcal{A}$ to sample an estimate $\mu_{a_{t}}$ for the expected return. The action with the highest estimate is selected.
The complete formulation is given in Algorithm \ref{algorithm:thompson_sampling}.

\begin{algorithm}
\caption{Generalized Thompson Sampling}\label{algorithm:thompson_sampling}
\begin{algorithmic}
\Procedure{$\textit{ThompsonSampling}(N_{t})$}{}
\For{$a_{t} \in \mathcal{A}$}
\State Infer $\langle \mu_{1},\lambda_{1},\alpha_{1},\beta_{1} \rangle$ from prior and $\overline{X_{a_{t}}},\sigma^{2}_{a}, n_{a_{t}}$
\State	$\mu_{a_{t}}, \tau_{a_{t}} \sim \mathcal{NG}(\mu_{1},\lambda_{1},\alpha_{1},\beta_{1})$
\EndFor
\Return $\textit{argmax}_{a_{t} \in \mathcal{A}}(\mu_{a_{t}})$
\EndProcedure
\end{algorithmic}
\begin{algorithmic}
\Procedure{$\textit{UpdateBandit}(N_{t}, G_{t})$}{}
\State $n_{a_{t}} \leftarrow n_{a_{t}} + 1$
\State $\langle \overline{X_{\textit{old},a_{t}}}, \overline{X_{a_{t}}} \rangle \leftarrow \langle\overline{X_{a_{t}}}, (n_{a_{t}}\overline{X_{\textit{old},a_{t}}} + G_{t})/(n_{a_{t}}+1)\rangle$
\State $s_{a_{t}} \leftarrow [(n_{a_{t}}-1)s_{a_{t}} + (G_{t} - \overline{X_{\textit{old},a_{t}}})(G_{t} - \overline{X_{a_{t}}})]/n_{a_{t}}$
\EndProcedure
\end{algorithmic}
\end{algorithm}

The prior should ideally reflect knowledge about the underlying model, especially for initial turns, where only a small amount of data has been observed \cite{honda2014optimality}. If no knowledge is available, then \emph{uninformative priors} should be chosen, where all possibilities can be sampled (almost) uniformly. This can be achieved by choosing the priors such that the variance of the resulting Normal distribution $\mathcal{N}(\mu_{0}, \frac{1}{\lambda_{0}\tau})$ becomes infinite ($\frac{1}{\lambda_{0}\tau_{0}} \rightarrow \infty$ and $\lambda_{0}\tau \rightarrow 0$). Since $\tau$ follows a Gamma distribution $\textit{Gamma}(\alpha_{0},\beta_{0})$ with expectation $\mathbb{E}(\tau) = \frac{\alpha_{0}}{\beta_{0}}$, $\alpha_{0}$ and $\beta_{0}$ should be chosen such that $\frac{\alpha_{0}}{\beta_{0}} \rightarrow 0$. Given the hyperparameter space $\lambda_{0} > 0$, $\alpha_{0} \geq 1$, and $\beta_{0} \geq 0$, it is recommended to set $\alpha_{0} = 1$ and $\mu_{0} = 0$ to center the Normal distribution. $\lambda_{0}$ should be chosen small enough and $\beta_{0}$ should have a sufficiently large value \cite{bai2014thompson}.

\subsubsection{Monte Carlo Belief State Update}
The belief state can be updated exactly according to Eq. \ref{eq:belief_update}. However, exact Bayes updates may be computationally infeasible in POMDPs with large state spaces due to the curse of dimensionality.
For this reason, we approximate the belief state $b_{h_{t}}$ for history $h_{t}$ with a particle filter as described in \cite{silver2010monte}. The belief state $b_{h_{t}}$ is represented by a set $\hat{b}_{h_{t}}$ of $K$ sample states or particles. After execution of $a_{t}$ and observation of $o_{t+1}$, the particles are updated by Monte Carlo simulation. Sampled states $s_{t} \sim \hat{b}_{h_{t}}$ are simulated with $a_{t}$ such that $\langle s_{t+1},o'_{t+1},r_{t} \rangle \sim \hat{M}(s_{t},a_{t})$. If $o'_{t+1} = o_{t+1}$, then $s_{t+1}$ is added to $\hat{b}_{h_{t}a_{t}o_{t+1}} = \hat{b}_{h_{t+1}}$.

\subsubsection{POOLTS}
To evaluate the effectiveness of POSTS compared to other open-loop planners, we first define \emph{Partially Observable Open-Loop Thompson Sampling (POOLTS)} and show that POOLTS is able to converge to an optimal open-loop plan, if sufficient computational and memory resources are provided. POOLTS is a tree-based approach based on OLUCT from \cite{lecarpentier2018open}. Each node $N_{t}$ represents a Thompson Sampling bandit and stores $\overline{X_{a_{t}}},s_{a}$, and $n_{a_{t}}$ for each action $a_{t} \in \mathcal{A}$.

A simulation starts at a state $s_{t}$, which is sampled from the current belief state $b_{h_{t}}$. The belief state is approximated by a particle filter $\hat{b}_{h_{t}}$ as described above.
An open-loop tree (Fig. \ref{fig:open_loop_planning}) is iteratively constructed by traversing the current tree in a selection step by using Thompson Sampling to select actions. When a leaf node is reached, it is expanded by a child node $N_{\textit{new}}$ and a rollout is performed by using a policy $\pi_{\textit{rollout}}$ until a terminal state is reached or a maximum search depth $T$ is exceeded. The observed rewards are accumulated to returns $G_{t}$ (Eq. \ref{eq:return}) and propagated back to update the corresponding bandit of every node in the simulated path. When the computation budget $n_{\textit{b}}$ has run out, the action $a_{t}$ with the highest expected return $\overline{X_{a_{t}}}$ is selected for execution. The complete formulation of POOLTS is given in Algorithm \ref{algorithm:poolts}.

\begin{algorithm}
\caption{POOLTS Planning}\label{algorithm:poolts}
\begin{algorithmic}
\Procedure{$\textit{POOLTS}(h_{t},T,n_{b})$}{}
\State Create $N_{0}$ for $h_{t}$
\While{$n_{b} > 0$}
\State $n_{b} \leftarrow n_{b} - 1$
\State $s_{t} \sim \hat{b}_{h_{t}}$
\State $\textit{TreeSearch}(N_{0},s_{t},T,0)$
\EndWhile
\State\Return $\textit{argmax}_{a_{t} \in \mathcal{A}}(\overline{X_{a_{t}}})$
\EndProcedure
\end{algorithmic}
\begin{algorithmic}
\Procedure{$\textit{TreeSearch}(N_{t},s_{t},T,d)$}{}
\If{$d \geq T$ or $s_{t}$ is a terminal state}
\Return 0
\EndIf
\If{$N_{t}$ is a leaf node}
\State Expand $N_{t}$
\State Perform rollout with $\pi_{\textit{rollout}}$ to sample $G_{t}$
\State\Return $G_{t}$
\EndIf
\State $a_{t} \leftarrow \textit{ThompsonSampling}(N_{t})$
\State $\langle s_{t+1}, r_{t},o_{t+1} \rangle \sim \hat{M}(s_{t}, a_{t})$
\State $R_{t} \leftarrow \textit{TreeSearch}(N_{t+1},s_{t+1},T,d+1)$
\State $G_{t} \leftarrow r_{t} + \gamma R_{t}$
\State $\textit{UpdateBandit}(N_{t}, G_{t})$
\State\Return $G_{t}$
\EndProcedure
\end{algorithmic}
\end{algorithm}

\cite{kocsis2006bandit,bubeck2010open,lecarpentier2018open} have shown that tree search algorithms using UCB1 converge to the optimal closed-loop or open-loop plan respectively, if the computation budget $n_{b}$ is sufficiently large. This is because the expected state or node values in the leaf nodes become stationary, given a stationary rollout policy $\pi_{\textit{rollout}}$. This enables the values in the preceding nodes to converge as well, leading to state- or node-wise optimal actions.
By replacing UCB1 with Thompson Sampling, the tree search should still converge to the optimal closed-loop or open-loop plan, since Thompson Sampling also converges to the optimal action, if the return distribution of $G_{t}$ becomes stationary \cite{agrawal2013further}. \cite{chapelle2011empirical,bai2014thompson} empirically demonstrated that Thompson Sampling converges faster than UCB1, when rewards are sparse and when the number of arms is large.

\subsubsection{POSTS}
\emph{Partially Observable Stacked Thompson Sampling (POSTS)} is an open-loop approach, which optimizes a stack of $T$ Thompson Sampling bandits to search for high-quality open-loop plans (Fig. \ref{fig:open_loop_planning_compressed}). Each bandit $N_{t}$ stores $\overline{X_{a_{t}}},s_{a}$, and $n_{a_{t}}$ for each action $a_{t}  \in \mathcal{A}$.

Similarly to POOLTS, a simulation starts at a state $s_{t}$, which is sampled from a particle filter $\hat{b}_{h_{t}}$, representing the current belief state $b_{h_{t}}$. Unlike POOLTS, a fixed size stack of bandits $N_{1},...,N_{T}$ is used to sample plans $p_{T} = [a_{1}, ..., a_{T}]$. $p_{T}$ is evaluated with the generative model $\hat{M}$ to observe immediate rewards $r_{1},...,r_{T}$, which are accumulated to returns $G_{1},...,G_{T}$ (Eq. \ref{eq:return}). Each bandit $N_{t}$ is then updated with the corresponding return $G_{t}$. When the computation budget $n_{\textit{b}}$ has run out, the action $a_{t}$ with the highest expected return $\overline{X_{a_{t}}}$ is selected for execution. The complete formulation of POSTS is given in Algorithm \ref{algorithm:posts}.

\begin{algorithm}
\caption{POSTS Planning}\label{algorithm:posts}
\begin{algorithmic}
\Procedure{$\textit{POSTS}(h_{t},T,n_{b})$}{}
\While{$n_{b} > 0$}
\State $n_{b} \leftarrow n_{b} - 1$
\State $s_{t} \sim \hat{b}_{h_{t}}$
\State $\textit{Simulate}(N_{0},s_{t},T,0)$
\EndWhile
\State\Return $\textit{argmax}_{a_{t} \in \mathcal{A}}(\overline{X_{a_{t}}})$
\EndProcedure
\end{algorithmic}
\begin{algorithmic}
\Procedure{$\textit{Simulate}(s_{t},T,d)$}{}
\If{$d \geq T$ or $s_{t}$ is a terminal state}
\Return 0
\EndIf
\State $a_{t} \leftarrow \textit{ThompsonSampling}(N_{t})$
\State $\langle s_{t+1}, r_{t},o_{t+1} \rangle \sim \hat{M}(s_{t}, a_{t})$
\State $R_{t} \leftarrow \textit{Simulate}(s_{t+1},T,d+1)$
\State $G_{t} \leftarrow r_{t} + \gamma R_{t}$
\State $\textit{UpdateBandit}(N_{t}, G_{t})$
\State\Return $G_{t}$
\EndProcedure
\end{algorithmic}
\end{algorithm}

The idea of POSTS is to only regard the temporal dependencies between the actions of an open-loop plan. The bandit stack is used to learn these dependencies with the expected (discounted) return. When a bandit $N_{t}$ samples an action $a_{t}$ with a resulting reward of $r_{t}$, then all preceding bandits $N_{t-k}$ with $k < t$ are updated with $G_{t-k} = r_{t-k} + \gamma r_{t-k+1} + ... + \gamma^{k} r_{t}$, using a discounted value of $r_{t}$. This is because the actions sampled by all preceding bandits are possibly relevant for obtaining the reward $r_{t}$. By only regarding these temporal dependencies, POSTS is \emph{memory bounded}, not requiring a search tree to model dependencies between histories or history distributions (Fig. \ref{fig:closed_loop_planning} and \ref{fig:open_loop_planning}).

\section{Experiments}
\subsection{Evaluation Environments}

We tested POSTS in the \textit{RockSample}, \textit{Battleship}, and \textit{PocMan} domains, which are well-known POMDP benchmark problems for decision making in POMDPs \cite{silver2010monte,somani2013despot,bai2014thompson}. For each domain, we set the discount factor $\gamma$ as proposed in \cite{silver2010monte}. The results were compared with POMCP, POOLTS, and a partially observable version of OLUCT, which we call POOLUCT. The problem-size features of all domains are shown in Table \ref{tab:benchmark_complexity}.

\begin{table*}
\centering
\caption{The problem-size features of the benchmark domains \textit{RockSample}, \textit{Battleship}, and \textit{PocMan}.}
\begin{tabular}{|P{3cm}|P{3cm}|P{3cm}|P{3cm}|P{3cm}|} \hline
 & \textit{RockSample(11,11)} & \textit{RockSample(15,15)} & \textit{Battleship} & \textit{PocMan}\\ \hline
\# States $|\mathcal{S}|$ & $247,808$ & $7,372,800$ & $\sim 10^{12}$ & $\sim 10^{56}$ \\
\# Actions $|\mathcal{A}|$ & $16$ & $20$ & $100$ & $4$ \\
\# Observations $|\mathcal{O}|$ & $3$ & $3$ & $2$ & $1024$ \\ \hline
\end{tabular}\label{tab:benchmark_complexity}
\end{table*}

The \textit{RockSample(n,k)} problem simulates an agent moving in an $n \times n$ grid containing $k$ rocks \cite{smith2004heuristic}. Each rock can be $good$ or $bad$ but the true state of each rock is unknown. The agent has to sample good rocks, while avoiding to sample bad rocks. It has a noisy sensor, which produces an observation $o_{t} \in \{good,bad\}$ for a particular rock. The probability of sensing the correct state of the rock decreases exponentially with the agent's distance to that rock. Sampling gives a reward of $+10$, if the rock is good and $-10$ otherwise. If a good rock was sampled, it becomes bad. Moving and sensing do not give any rewards. Moving past the east edge of the grid gives a reward of $+10$ and the episode terminates. We set $\gamma = 0.95$.

In \textit{Battleship} five ships of size 1, 2, 3, 4, and 5 respectively are randomly placed into a $10 \times 10$ grid, where the agent has to sink all ships without knowing their actual positions \cite{silver2010monte}. Each cell hitting a ship gives a reward of $+1$. There is a reward of $-1$ per time step and a terminal reward of $+100$ for hitting all ships. We set $\gamma = 1$.

\textit{PocMan} is a partially observable version of \textit{PacMan} \cite{silver2010monte}. The agent navigates in a $17 \times 19$ maze and has to eat randomly distributed food pellets and power pills. There are four ghosts moving randomly in the maze. If the agent is within the visible range of a ghost, it is getting chased by the ghost and dies, if it touches the ghost, terminating the episode with a reward of $-100$. Eating a power pill enables the agent to eat ghosts for 15 time steps. In that case, the ghosts will run away, if the agent is under the effect of a power pill. At each time step a reward of $-1$ is given. Eating food pellets gives a reward of $+10$ and eating a ghost gives $+25$. The agent can only perceive ghosts, if they are in its direct line of sight in each cardinal direction or within a hearing range. Also, the agent can only sense walls and food pellets, which are adjacent to it. We set $\gamma = 0.95$.

\subsection{Methods}


\subsubsection{POMCP}
We use the POMCP implementation from \cite{silver2010monte}. $\pi_{\textit{tree}}$ selects actions from a set of legal actions $a_{t} \in \mathcal{A}_{\textit{legal}}(s_{t})$ with UCB1. $\pi_{\textit{rollout}}$ randomly selects actions from $\mathcal{A}_{\textit{legal}}(s')$, depending on the currently simulated state $s' \in \mathcal{S}$.

In each simulation step, there is at most one expansion step, where new nodes are added to the search tree. Thus, tree size should increase linearly w.r.t. $n_{b}$ in large POMDPs.

\subsubsection{POOLTS and POOLUCT}
POOLTS is implemented according to Algorithm \ref{algorithm:poolts}, where actions are selected from a set of legal actions $a_{t} \in \mathcal{A}_{\textit{legal}}(s_{t})$ with Thompson Sampling (Algorithm \ref{algorithm:thompson_sampling}) in the first stage. $\pi_{\textit{rollout}}$ randomly selects actions from $\mathcal{A}_{\textit{legal}}(s')$, depending on the currently simulated state $s' \in \mathcal{S}$.
POOLUCT is similar to POOLTS but uses UCB1 as action selection strategy in the first stage. Since, open-loop planning can encounter different states at the same node (Fig. \ref{fig:closed_vs_open_loop_planning}), the set of legal actions may vary for each state $s_{t} \in \mathcal{S}$. We always mask out the statistics of currently illegal actions, regardless of whether they have high average action values, to avoid selecting them.

Similarly to POMCP, the search tree size should increase linearly w.r.t. $n_{b}$, but with less nodes, since open-loop trees store summarized information about history distributions.

\subsubsection{POSTS}
POSTS is implemented as a stack of Thompson Sampling bandits $N_{i}$ with $1 \leq i \leq T$ according to Algorithm \ref{algorithm:posts}. Starting at $s_{t}$, all bandits $N_{i}$ apply Thompson Sampling to a set of legal actions $\mathcal{A}_{\textit{legal}}(s')$, depending on the currently simulated state $s' \in \mathcal{S}$. Similarly to POOLTS and POOLUCT, we mask out the statistics of currently illegal actions and only regard the value statistics of legal actions for selection during planning.

Given a horizon of $T$, POSTS always maintains $T$ Thompson Sampling bandits, independently of the computation budget $n_{b}$.

\subsection{Results}
We ran each approach on \textit{RockSample}, \textit{Battleship}, and \textit{PocMan} with different settings for 100 times or at most 12 hours of total computation. We evaluated the performance of each approach with the \emph{undiscounted return} ($\gamma = 1$), because we focus on the actual effectiveness instead of the quality of optimization \cite{bai2014thompson}. For POMCP and POOLTS we set the UCB1 exploration constant $c$ to the reward range of each domain as proposed in \cite{silver2010monte}.

\subsubsection{Prior Sensitivity}
\begin{figure}[!ht]
     \subfloat[\textit{RockSample(11,11)}\label{fig:rocksample_11_horizon_100_prior}]{%
       \includegraphics[width=0.23\textwidth]{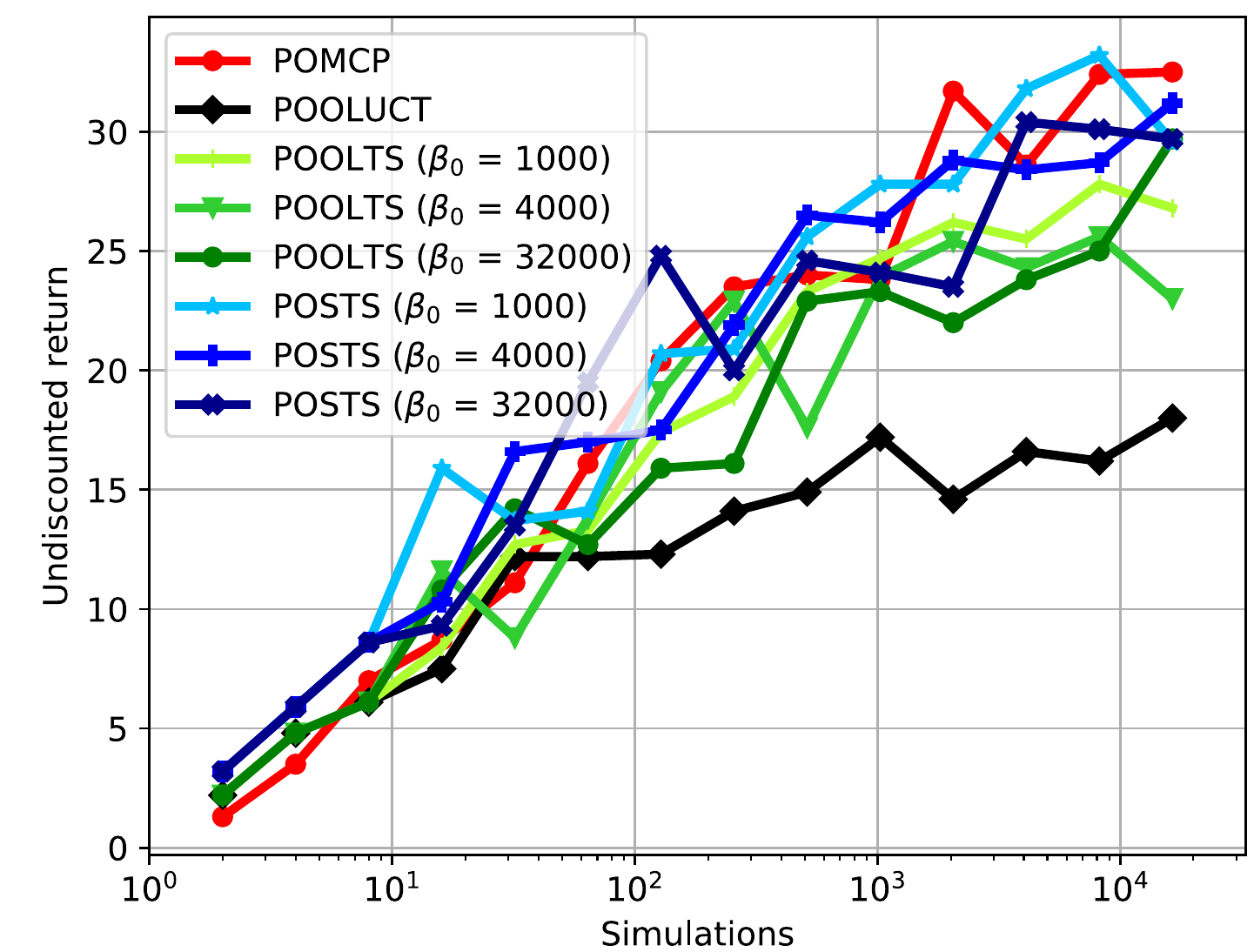}
     }
     \hfill
     \subfloat[\textit{RockSample(15,15)}\label{fig:rocksample_15_horizon_100_prior}]{%
       \includegraphics[width=0.23\textwidth]{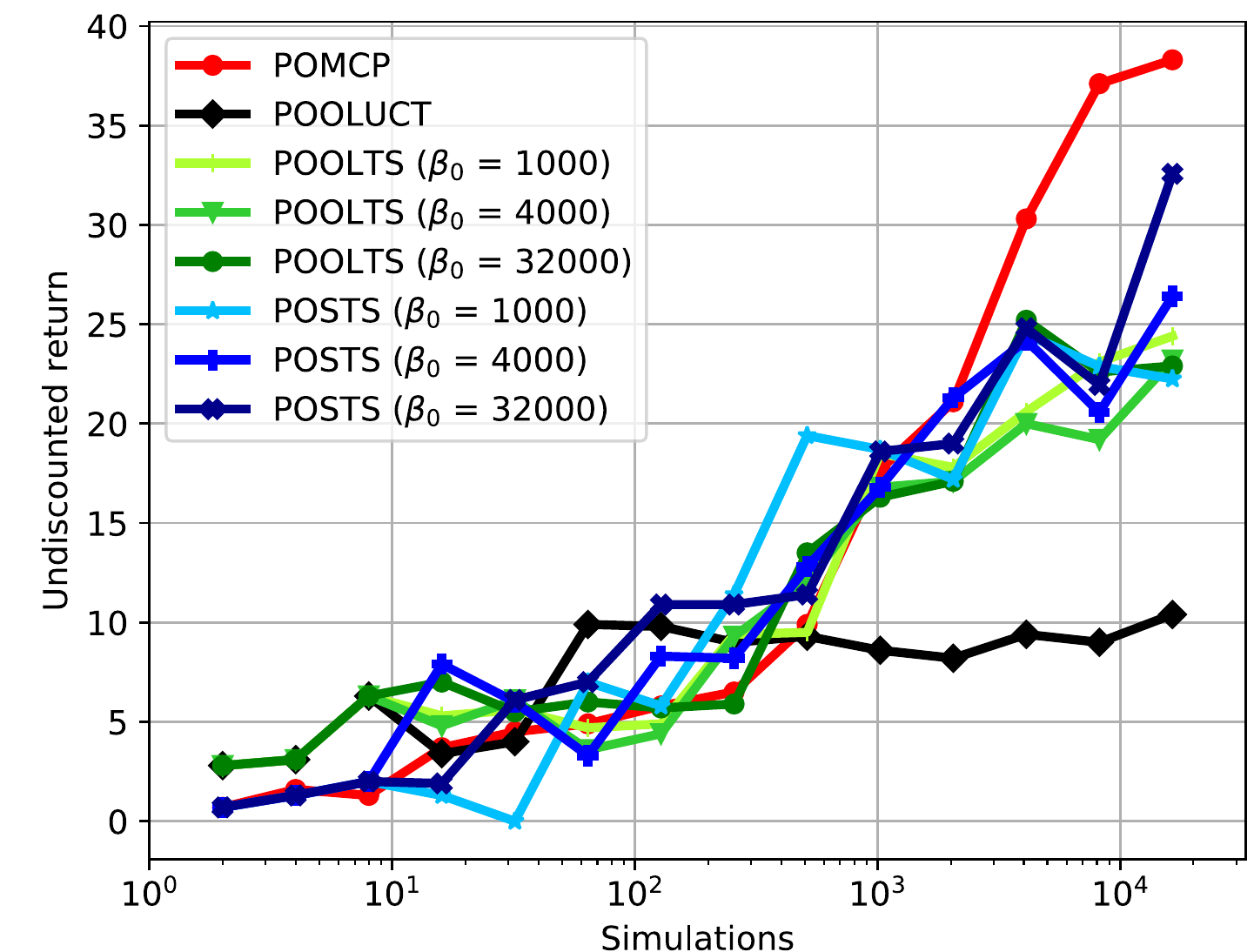}
     }
     \\
     \subfloat[\textit{Battleship}\label{fig:battleship_horizon_100_prior}]{%
       \includegraphics[width=0.23\textwidth]{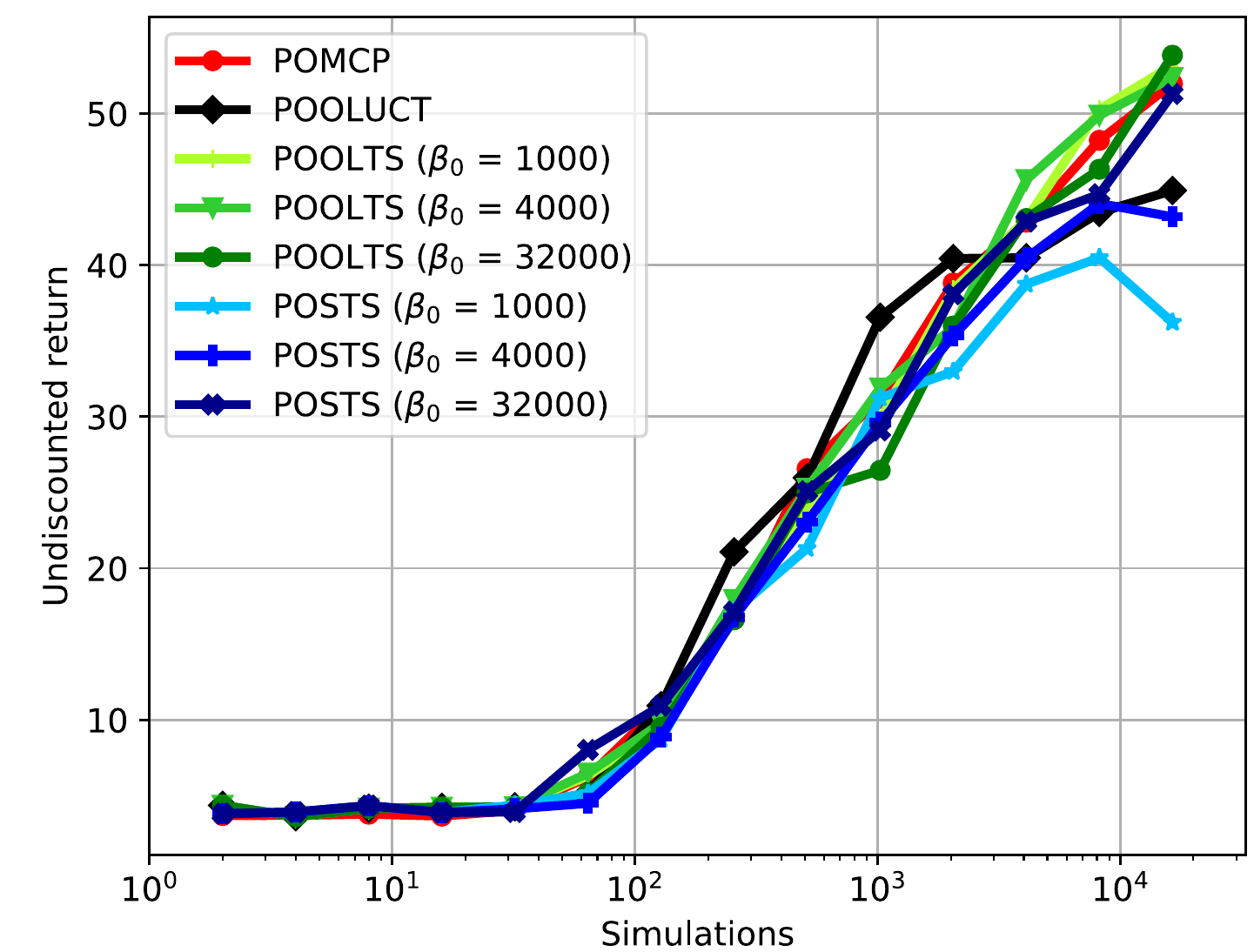}
     }
     \hfill
     \subfloat[\textit{PocMan}\label{fig:pocman_horizon_100_prior}]{%
       \includegraphics[width=0.23\textwidth]{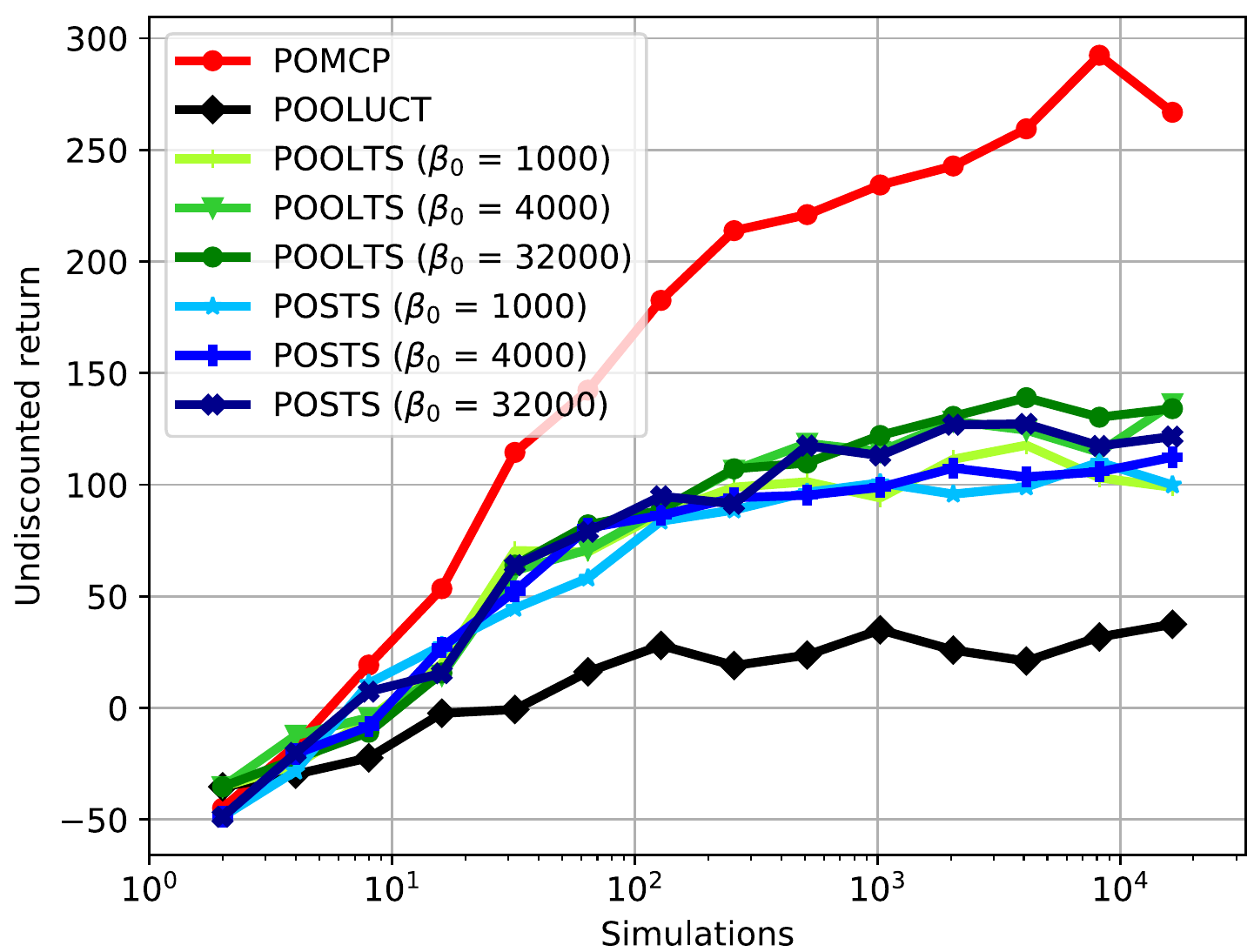}
     }
     \caption{Average performance of POSTS, POOLTS, POOLUCT, and POMCP with different prior values for $\beta_{0}$, different computation budgets $n_{b}$, and a horizon of $T = 100$.}
     \label{fig:posts_prior_sensitivity}
\end{figure}

Since we assume no additional domain knowledge, we focus on uninformative priors with $\mu_{0} = 0$, $\alpha_{0} = 1$, and $\lambda_{0} = 0.01$ as proposed in \cite{bai2014thompson}. With this setting, $\beta_{0}$ controls the degree of initial exploration during the planning phase, thus its impact on the performance of POOLTS and POSTS is evaluated. The results are shown in Fig. \ref{fig:posts_prior_sensitivity} for $\beta_{0} = 1000,4000,32000$ for POOLTS and POSTS.

In \textit{RockSample}, POSTS slightly outperforms POOLTS and keeps up in performance with POMCP. POOLTS slightly outperforms POSTS and POMCP in \textit{Battleship} with POSTS only being able to keep up when $n_{b} < 10^{4}$ or when $\beta_{0} = 32000$. POMCP clearly outperforms all open-loop approaches in \textit{PocMan}. POOLTS slightly outperforms POSTS in \textit{PocMan} with POSTS only being to keep up, if $\beta_{0} = 32000$. POOLUCT performed worst in all domains except \textit{Battleship}, where it performs best with a computation budget of $n_{b} \leq 1024$. POSTS performs slightly better, if $\beta_{0}$ is large, but POOLTS seem to be insensitive to the choice of $\beta_{0}$ except in \textit{PocMan}, where it performs better, if $\beta_{0}$ is large.

\subsubsection{Horizon Sensitivity}
\begin{figure}[!ht]
     \subfloat[\textit{RockSample(11,11)}\label{fig:rocksample_11_horizon}]{%
       \includegraphics[width=0.23\textwidth]{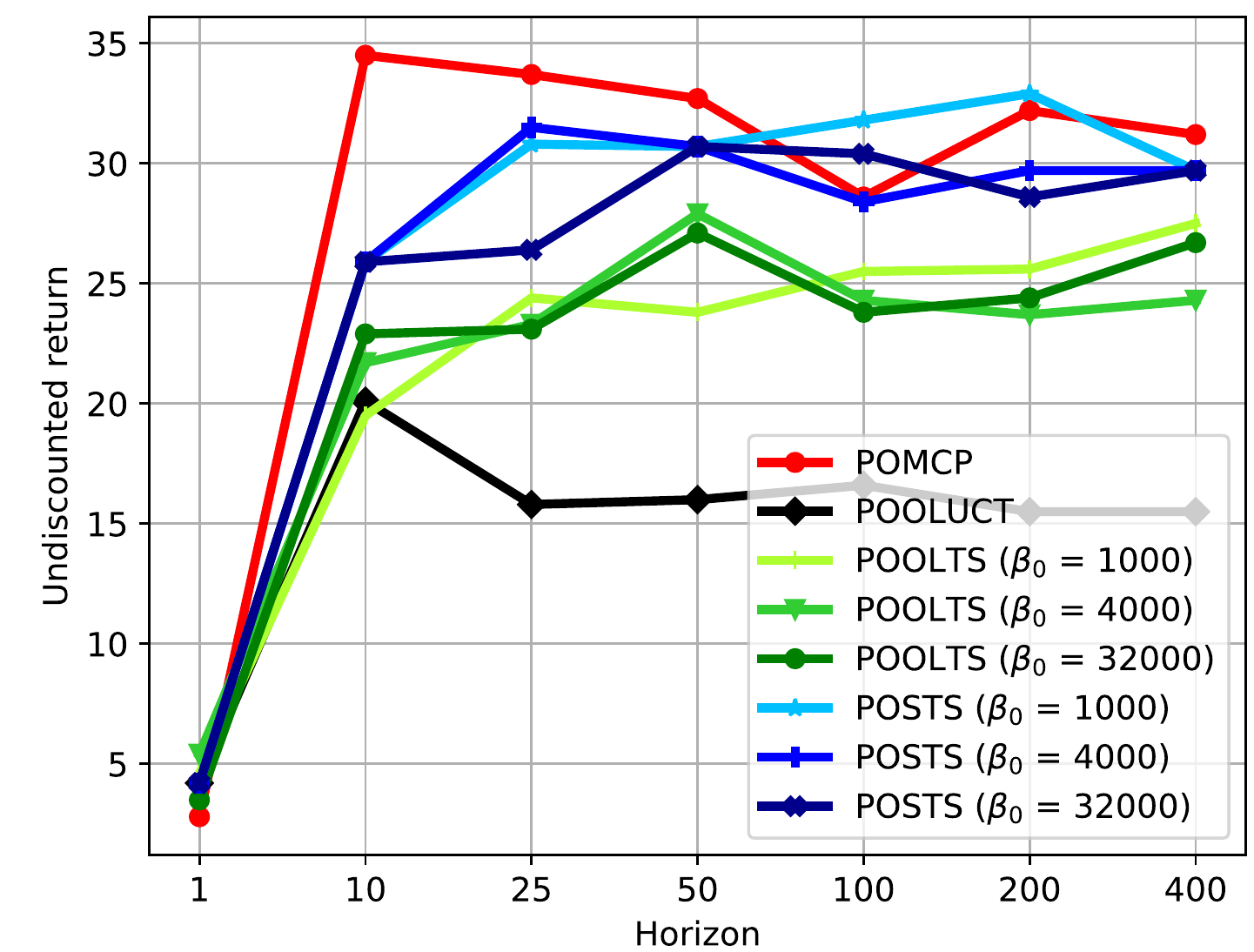}
     }
     \hfill
     \subfloat[\textit{RockSample(15,15)}\label{fig:rocksample_15_horizon}]{%
       \includegraphics[width=0.23\textwidth]{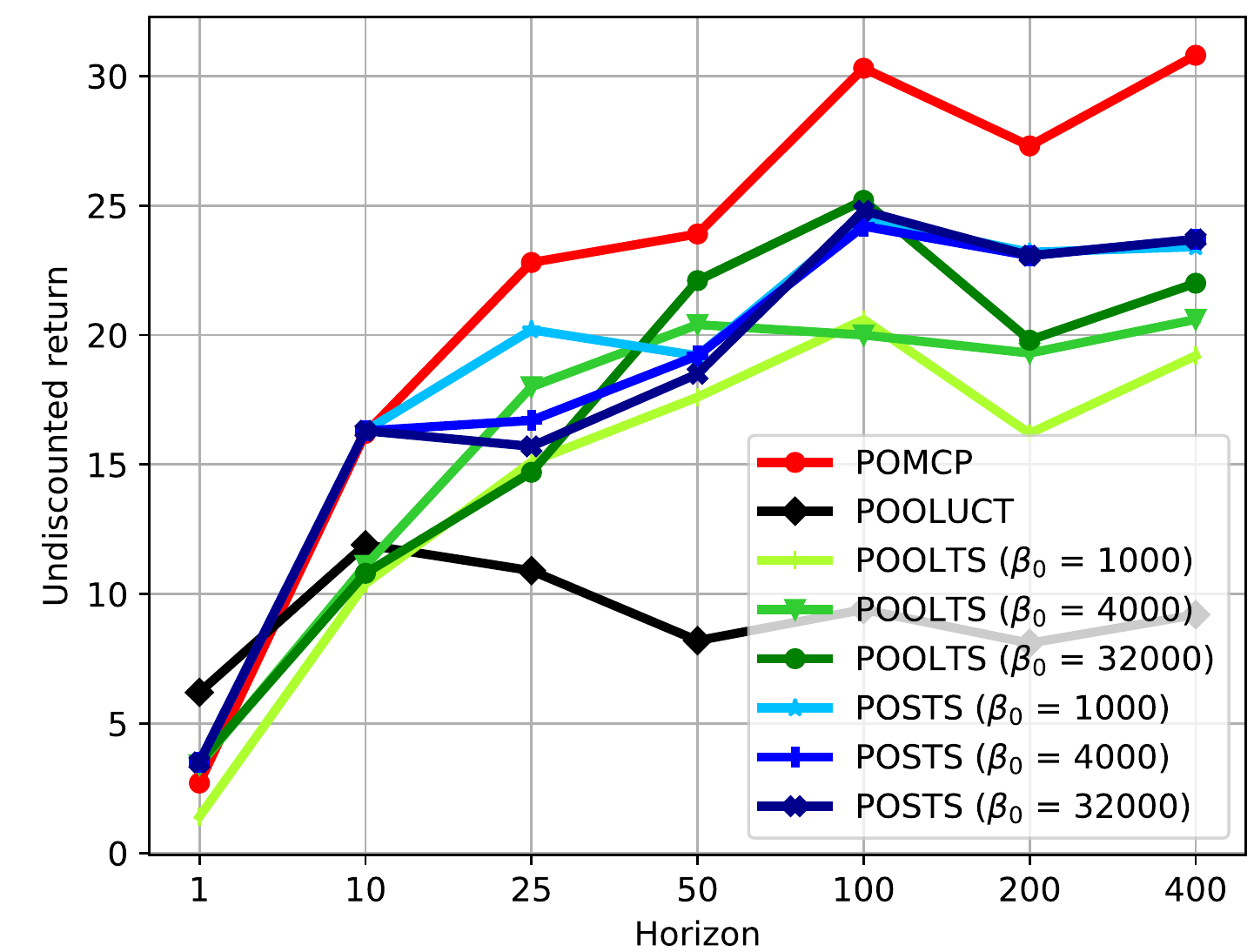}
     }
     \\
     \subfloat[\textit{Battleship}\label{fig:battleship_horizon}]{%
       \includegraphics[width=0.23\textwidth]{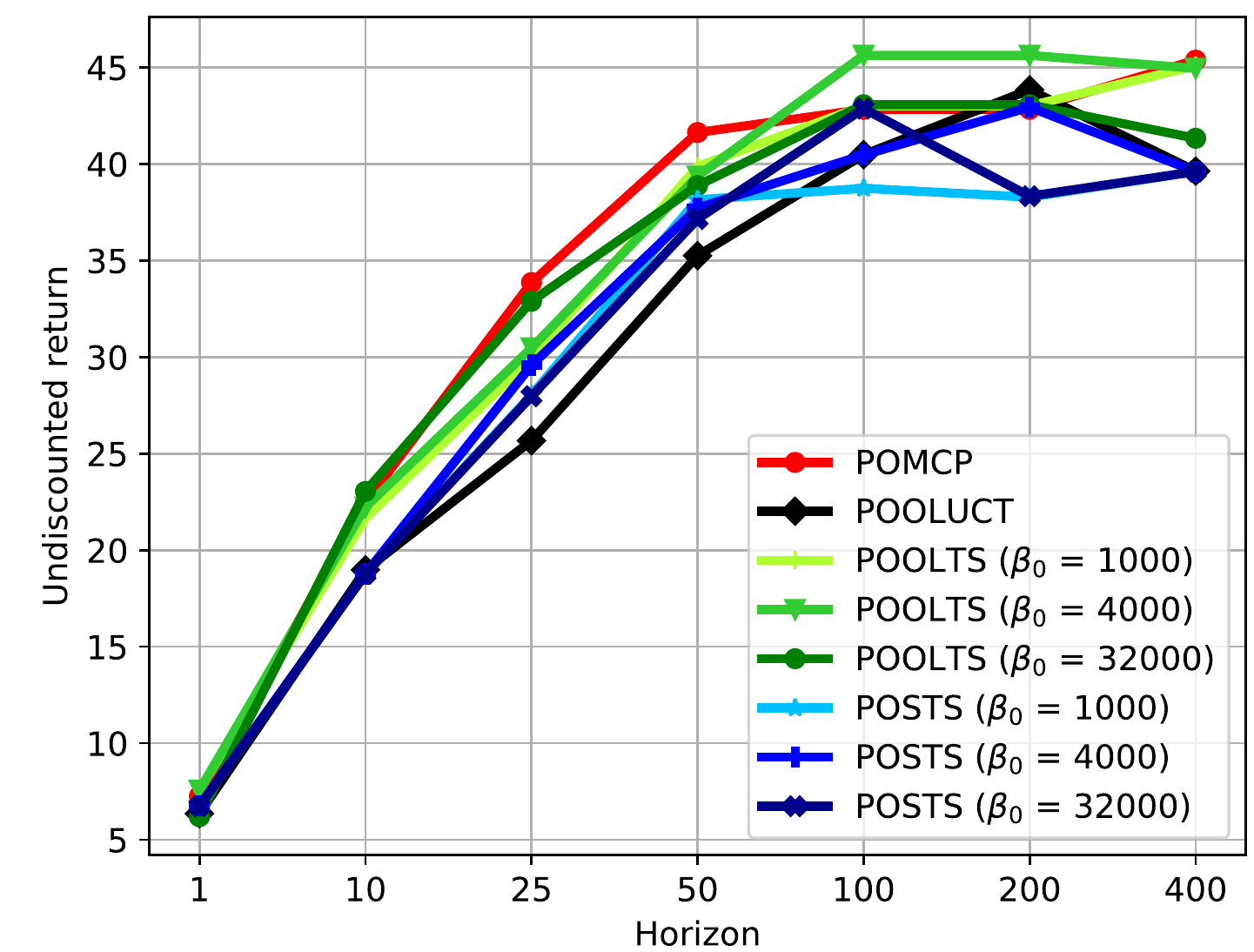}
     }
     \hfill
     \subfloat[\textit{PocMan}\label{fig:pocman_horizon}]{%
       \includegraphics[width=0.23\textwidth]{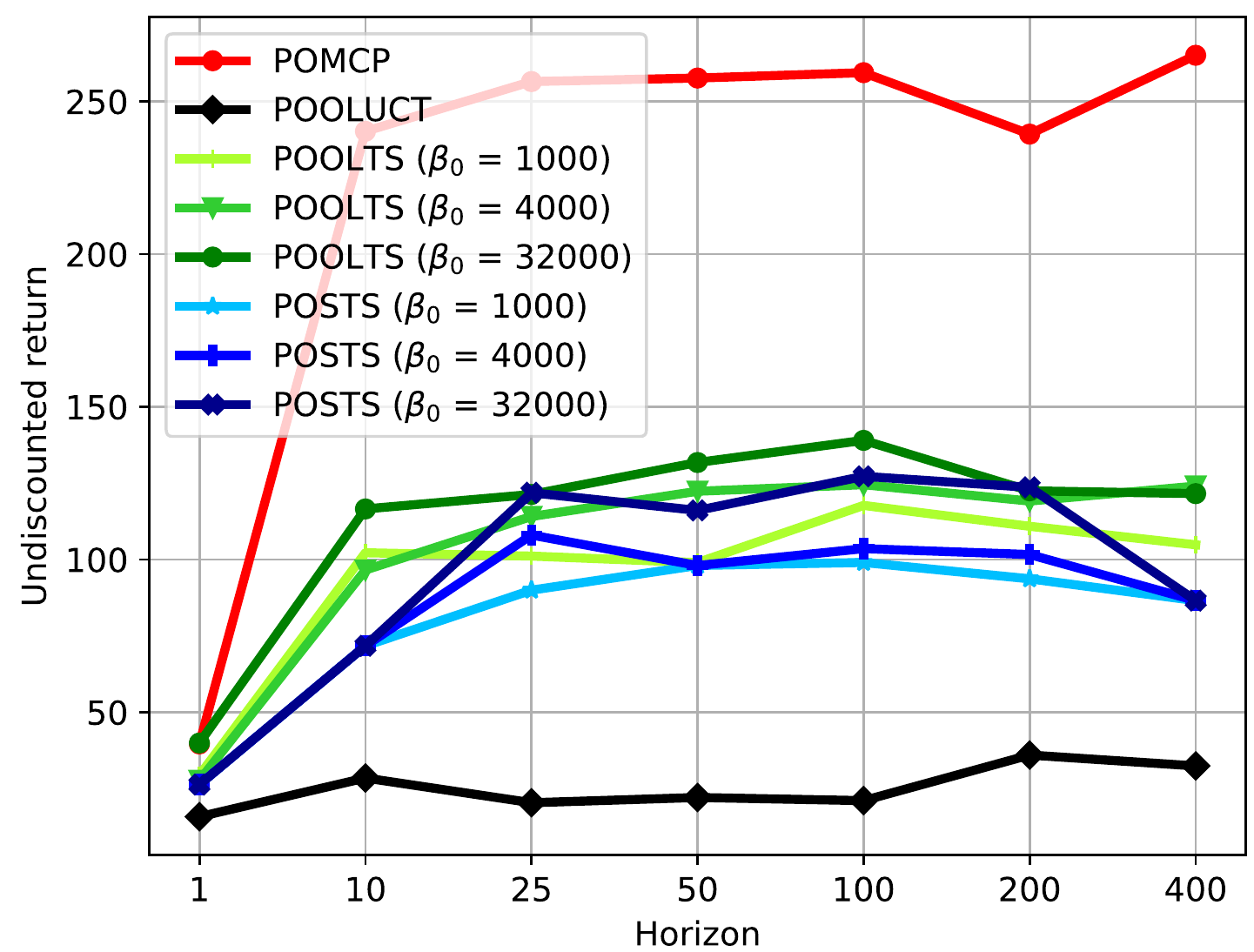}
     }
     \caption{Average performance of POSTS, POOLTS, POOLUCT, and POMCP with different planning horizons $T$ and a computation budget of $n_{b} = 4096$.}
     \label{fig:posts_horizon_sensitivity}
\end{figure}

We evaluated the sensitivity of all approaches w.r.t. different horizons $T$. The results are shown in Fig. \ref{fig:posts_horizon_sensitivity} for $n_{b} = 4096$ \footnote{Using computation budgets between 1024 and 16384 led to similar plots, thus we stick to $n_{b} = 4096$ with all approaches requiring less than one second per action \cite{silver2010monte}.} and $\beta_{0} = 1000,4000,32000$ for POOLTS and POSTS.

In \textit{RockSample(11,11)}, there is a performance peak at $T = 10$ for POMCP and POOLUCT, while for POSTS and POOLTS it is about $T = 50$. In all other domains, there seems to be a performance peak at $T = 100$ for most approaches. If $T > 100$, there is no significant improvement or even degrading performance for most approaches except for POMCP, which slightly improves in all domains but \textit{RockSample(11,11)}, if $T = 400$.

\subsubsection{Performance-Memory Tradeoff}
\begin{figure}[!ht]
     \subfloat[\textit{RockSample(11,11)}\label{fig:rocksample_11_nodes}]{%
       \includegraphics[width=0.23\textwidth]{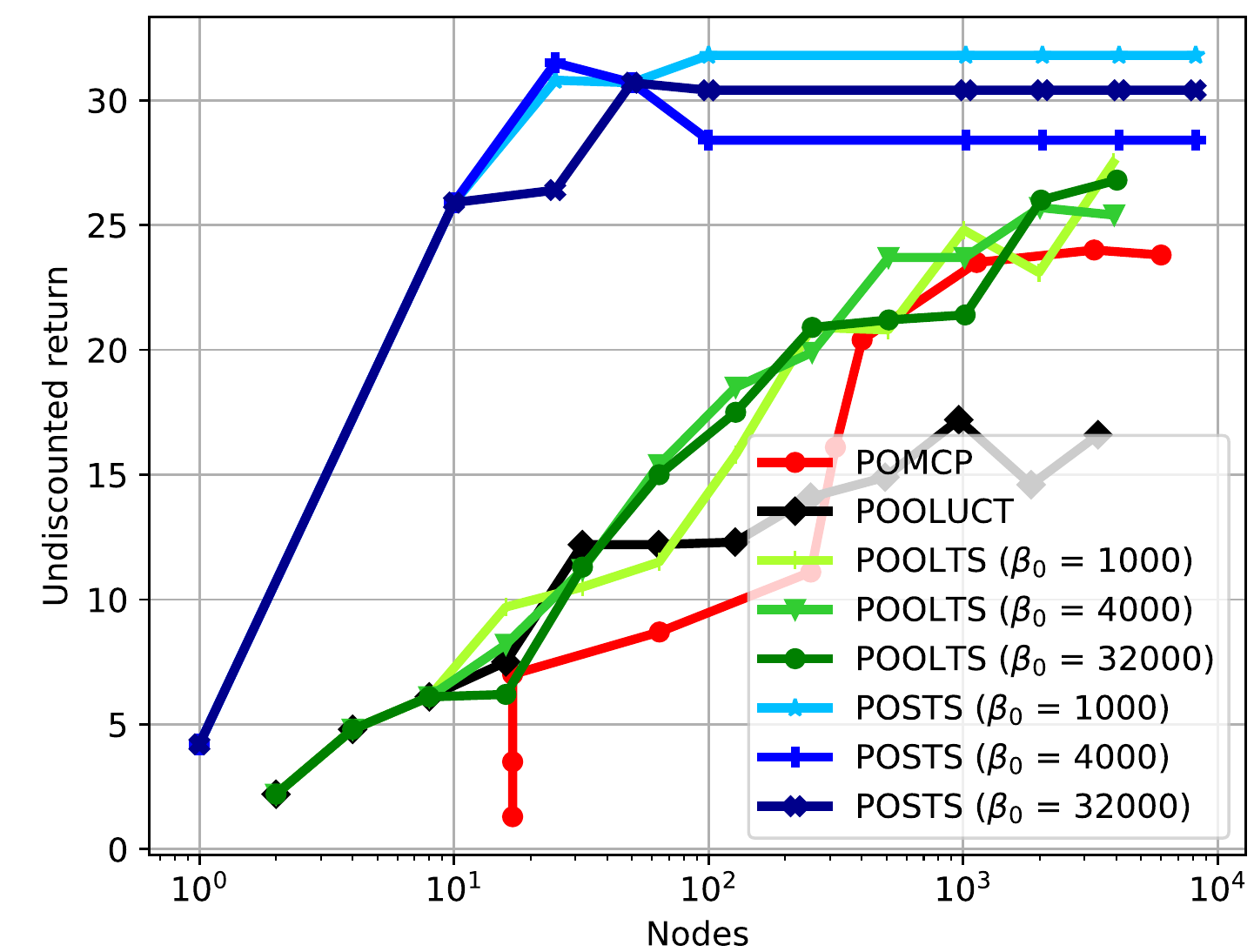}
     }
     \hfill
     \subfloat[\textit{RockSample(15,15)}\label{fig:rocksample_15_nodes}]{%
       \includegraphics[width=0.23\textwidth]{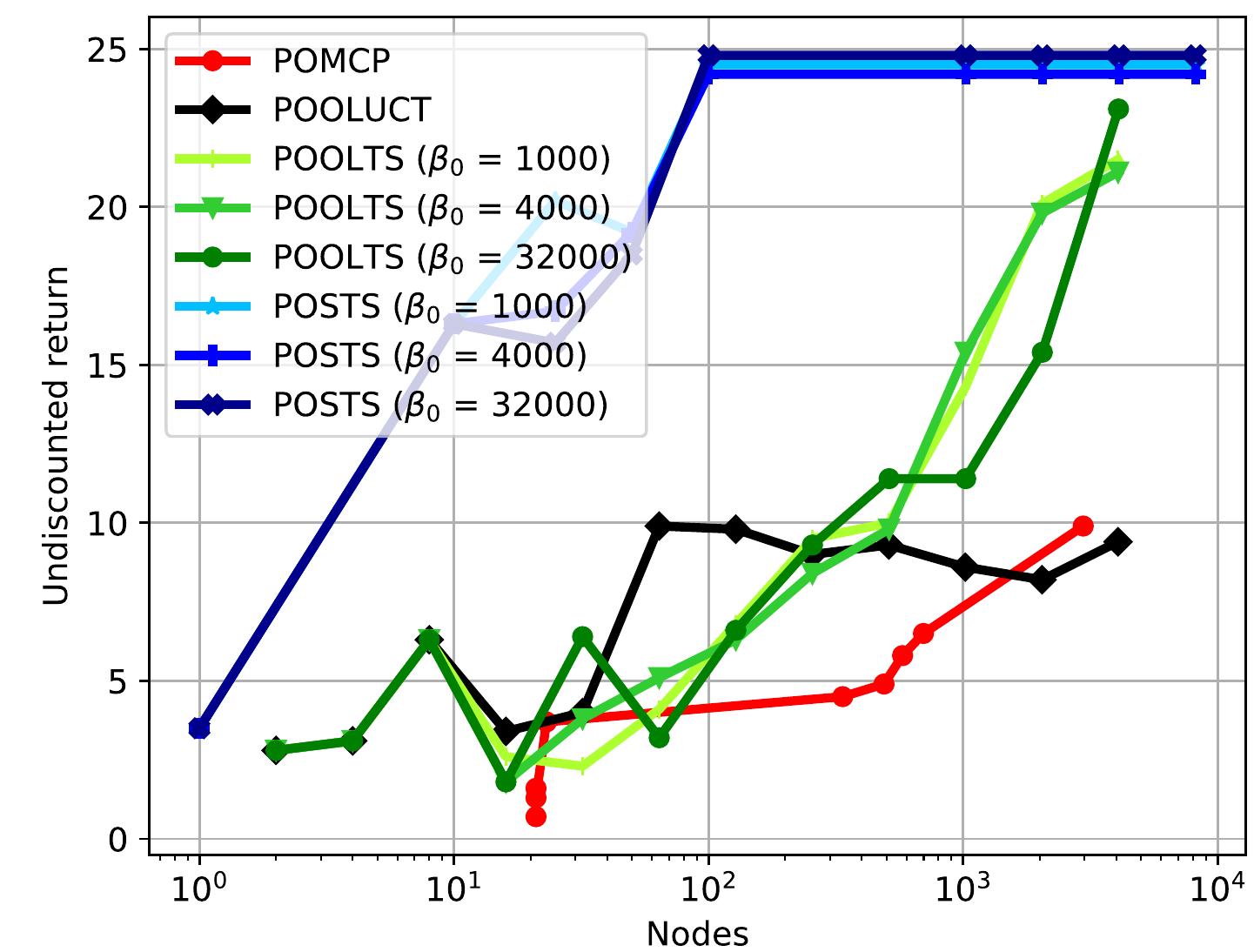}
     }
     \\
     \subfloat[\textit{Battleship}\label{fig:battleship_nodes}]{%
       \includegraphics[width=0.23\textwidth]{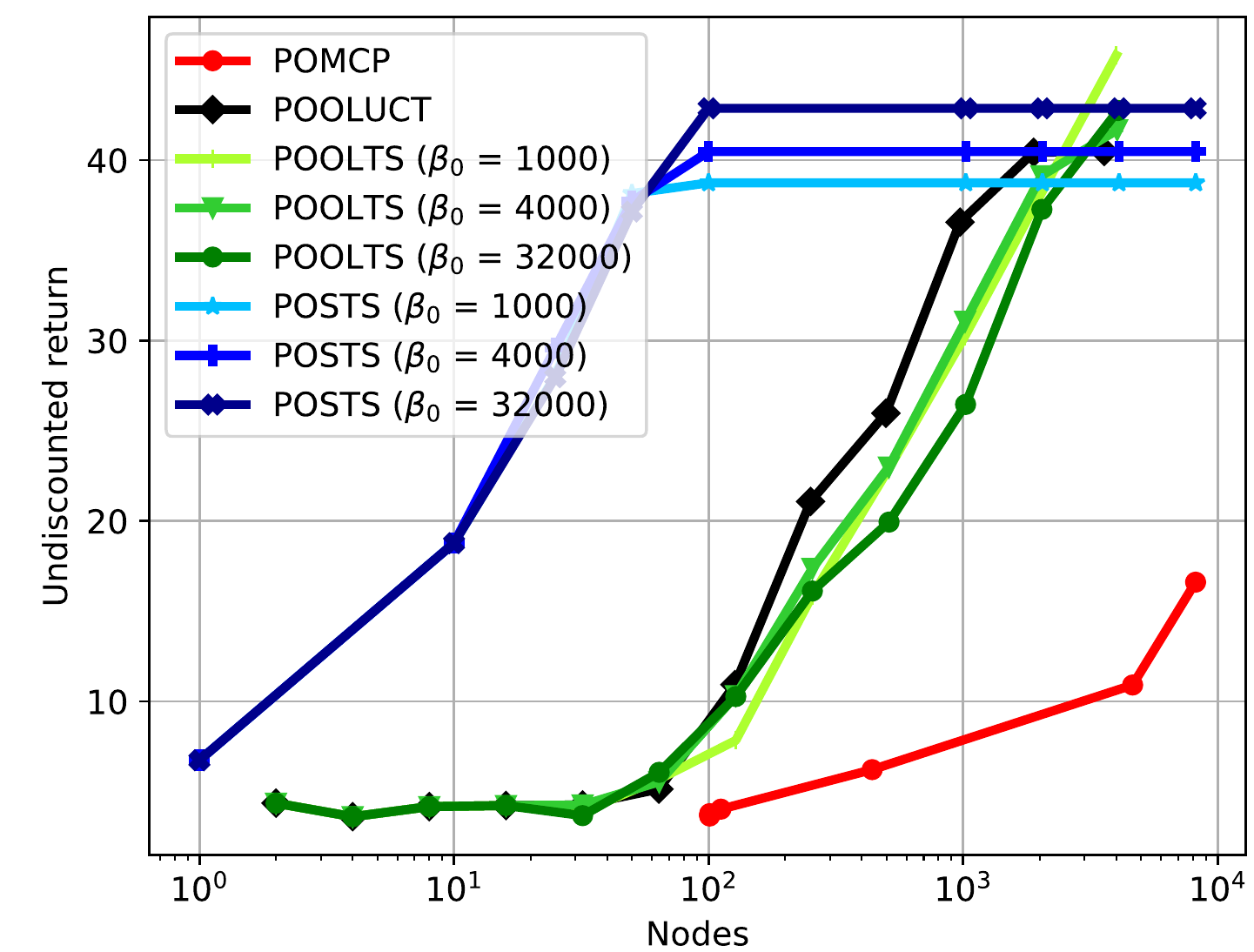}
     }
     \hfill
     \subfloat[\textit{PocMan}\label{fig:pocman_nodes}]{%
       \includegraphics[width=0.23\textwidth]{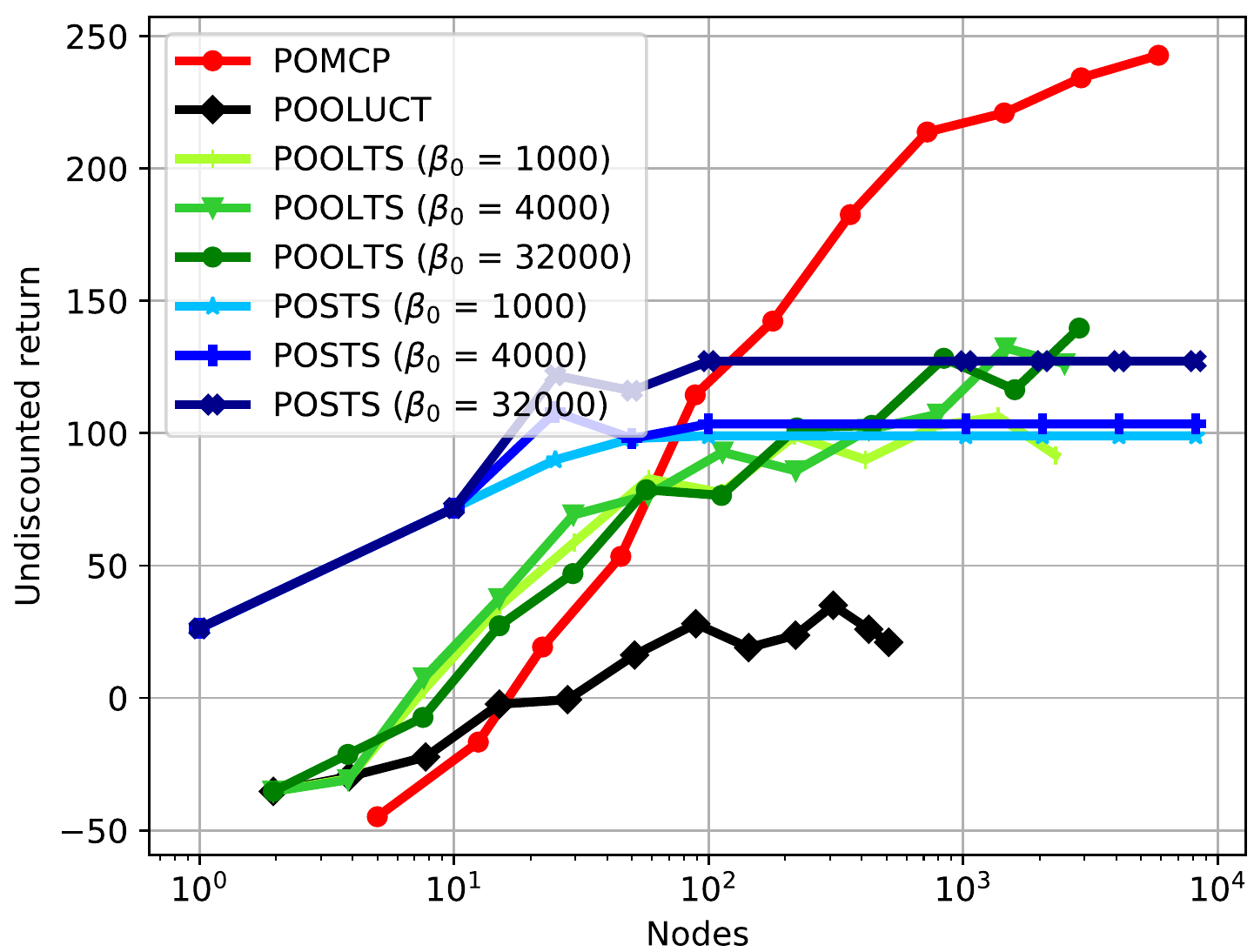}
     }
     \caption{Average performance of POSTS, POOLTS, POOLUCT, and POMCP with memory bounds, a computation budget of $n_{b} = 4096$ and a horizon of $T = 100$.}
     \label{fig:posts_performance_memory_tradeoff}
\end{figure}

We evaluated the performance-memory tradeoff of all approaches by introducing a memory capacity $n_{\textit{mem}}$, where the computation is interrupted, when the number of nodes exceeds $n_{\textit{mem}}$. For POMCP, we count the number of \textit{o-nodes} and \textit{a-nodes} (Fig. \ref{fig:closed_loop_planning}). For POOLTS and POOLUCT, we count the number of history distribution nodes (Fig. \ref{fig:open_loop_planning}). For POSTS, we count the number of Thompson Sampling bandits, which is always $\textit{min}\{n_{\textit{mem}}, T\}$. The results are shown in Fig. \ref{fig:posts_performance_memory_tradeoff} for $n_{b} = 4096$, $T = 100$, and $\beta_{0} = 1000,4000,32000$ for POOLTS and POSTS. POSTS never uses more than $10^{2} = 100$ nodes in each setting.

In \textit{Rocksample} and \textit{Battleship}, POMCP is outperformed by POSTS and POOLTS (and also POOLUCT in \textit{Battleship}). POSTS always performs best in these domains, when $n_{\textit{mem}} < 1000$. POMCP performs best in \textit{PocMan} by outperforming POSTS, when $n_{\textit{mem}} > 100$ and POOLTS keeps up with the best POSTS setting, when $n_{\textit{mem}} > 1000$. POOLUCT performs worst except in \textit{Battleship}, improving less and slowest with increasing $n_{\textit{mem}}$. It outperforms POMCP in \textit{Rocksample(15,15)}, when $n_{\textit{mem}} \leq 1000$ though. In \textit{PocMan}, POOLUCT creates less than 550 nodes, when $n_{b} = 4096$, indicating that the search tree construction has converged and does not improve any further.

\subsection{Discussion}
The experiments show that partially observable open-loop planning can be a good alternative to closed-loop planning, when the action space is large, stochasticity is low, and when computational and memory resources are highly restricted. Especially approaches based on Thompson Sampling like POOLTS and POSTS seem to be very effective and robust w.r.t. the hyperparameter choice. Setting a large value for $\beta_{0}$ seems to be beneficial for large problems (Fig. \ref{fig:posts_prior_sensitivity}). This is because an enormous search space needs to be explored, while avoiding premature convergence to poor solutions. However, if $\beta_{0}$ is too large, POSTS and POOLTS might converge too slowly, thus requiring much more computation \cite{bai2014thompson}. If $T$ is too large, then the value estimates have very high variance, making bandit adaptation more difficult. This could explain the performance stagnation or degradation for most approaches in Fig. \ref{fig:posts_horizon_sensitivity}, when $T > 100$. The performance of POSTS scales similarly to POOLTS w.r.t. $n_{b}$ and $T$ (Fig. \ref{fig:posts_prior_sensitivity} and \ref{fig:posts_horizon_sensitivity}). POSTS is also more robust than POOLUCT w.r.t. changes to $n_{b}$ and $T$ except in \textit{Battleship}, where both approaches scale similarly, when $\beta_{0}$ is sufficiently large.

POSTS is competitive to POOLTS and superior to POOLUCT in all settings except in \textit{Battleship} (when $n_{b}$ is large) with POOLTS and POOLUCT being shown to theoretically converge to optimal open-loop plans, given sufficient computation budget $n_{b}$ and memory capacity $n_{\textit{mem}}$. POSTS is shown to be superior to all other approaches in \textit{RockSample} and \textit{Battleship}, when memory resources are highly restricted, only being outperformed by the tree-based approaches in \textit{Battleship} after thousands of nodes were created, consuming much more memory than POSTS, which only uses 100 nodes at most. This might be due to the relatively large action space of these domains (Table \ref{tab:benchmark_complexity}), where all tree-based planners construct enormous trees with high branching factors, when exploring the effect of each action. \textit{RockSample} and \textit{Battleship} have low stochasticity, since state transitions are deterministic. In both domains the agent is primarily uncertain about the real state, thus the planning quality only depends on the belief state approximation and the uncertainty about observations (only in \textit{RockSample}).

POMCP performs best in \textit{PocMan}. This might be due to the small action space (Table \ref{tab:benchmark_complexity}) and high stochasticity (where all ghosts primarily move randomly), since open-loop planning is known to converge to sub-optimal solutions in such domains \cite{weinstein2013open,lecarpentier2018open}. However, POMCP has the highest memory consumption, since it constructs larger trees than open-loop approaches with the same computation budget (Fig \ref{fig:closed_loop_planning}). In \textit{PocMan}, POSTS is able to keep up with POOLTS, while being much more memory-efficient (Fig. \ref{fig:pocman_horizon_100_prior} and \ref{fig:pocman_nodes}).

\section{Conclusion and Future Work}
In this paper, we proposed \emph{Partially Observable Stacked Thompson Sampling (POSTS)}, a memory bounded approach to open-loop planning in large POMDPs, which optimizes a fixed size stack of Thompson Sampling bandits.

To evaluate the effectiveness of POSTS, we formulated a tree-based approach, called POOLTS and showed that POOLTS is able to find optimal open-loop plans with sufficient computational and memory resources.

We empirically tested POSTS in four large benchmark problems and showed that POSTS achieves competitive performance compared to tree-based open-loop planners like POOLTS and POOLUCT, if sufficient resources are provided. Unlike tree-based approaches, POSTS offers a performance-memory tradeoff by performing best, if computational and memory resources are highly restricted, making it suitable for efficient partially observable planning.

For the future, we plan to apply POSTS to conformant planning problems \cite{hoffmanna2006conformant,palacios2009compiling,geffner2013concise} and to extend it to multi-agent settings \cite{phan2018evade}.

\bibliographystyle{aaai}
\bibliography{posts}

\begin{thebibliography}{}

\bibitem[\protect\citeauthoryear{Agrawal and Goyal}{2013}]{agrawal2013further}
Agrawal, S., and Goyal, N.
\newblock 2013.
\newblock {Further} {Optimal} {Regret} {Bounds} for {Thompson} {Sampling}.
\newblock In {\em Artificial Intelligence and Statistics},  99--107.

\bibitem[\protect\citeauthoryear{Auer, Cesa-Bianchi, and
  Fischer}{2002}]{auer2002finite}
Auer, P.; Cesa-Bianchi, N.; and Fischer, P.
\newblock 2002.
\newblock {Finite}-{Time} {Analysis} of the {Multiarmed} {Bandit} {Problem}.
\newblock {\em Machine learning} 47(2-3):235--256.

\bibitem[\protect\citeauthoryear{Bai \bgroup et al\mbox.\egroup
  }{2014}]{bai2014thompson}
Bai, A.; Wu, F.; Zhang, Z.; and Chen, X.
\newblock 2014.
\newblock {Thompson} {Sampling} based {Monte}-{Carlo} {Planning} in {POMDPs}.
\newblock In {\em Proceedings of the Twenty-Fourth International Conferenc on
  International Conference on Automated Planning and Scheduling},  29--37.
\newblock AAAI Press.

\bibitem[\protect\citeauthoryear{Bai, Wu, and Chen}{2013}]{bai2013bayesian}
Bai, A.; Wu, F.; and Chen, X.
\newblock 2013.
\newblock {Bayesian} {Mixture} {Modelling} and {Inference} based {Thompson}
  {Sampling} in {Monte}-{Carlo} {Tree} {Search}.
\newblock In {\em Advances in Neural Information Processing Systems},
  1646--1654.

\bibitem[\protect\citeauthoryear{Belzner and Gabor}{2017}]{belzner2017stacked}
Belzner, L., and Gabor, T.
\newblock 2017.
\newblock {Stacked} {Thompson} {Bandits}.
\newblock In {\em Proceedings of the 3rd International Workshop on Software
  Engineering for Smart Cyber-Physical Systems},  18--21.
\newblock IEEE Press.

\bibitem[\protect\citeauthoryear{Bubeck and Munos}{2010}]{bubeck2010open}
Bubeck, S., and Munos, R.
\newblock 2010.
\newblock {Open} {Loop} {Optimistic} {Planning}.
\newblock In {\em COLT},  477--489.

\bibitem[\protect\citeauthoryear{Chapelle and Li}{2011}]{chapelle2011empirical}
Chapelle, O., and Li, L.
\newblock 2011.
\newblock {An} {Empirical} {Evaluation} of {Thompson} {Sampling}.
\newblock In {\em Advances in neural information processing systems},
  2249--2257.

\bibitem[\protect\citeauthoryear{Geffner and Bonet}{2013}]{geffner2013concise}
Geffner, H., and Bonet, B.
\newblock 2013.
\newblock {A} {Concise} {Introduction} to {Models} and {Methods} for
  {Automated} {Planning}.
\newblock {\em Synthesis Lectures on Artificial Intelligence and Machine
  Learning} 8(1):1--141.

\bibitem[\protect\citeauthoryear{Hoffmanna and
  Brafmanb}{2006}]{hoffmanna2006conformant}
Hoffmanna, J., and Brafmanb, R.~I.
\newblock 2006.
\newblock {Conformant} {Planning} via {Heuristic} {Forward} {Search}: {A} {New}
  {Approach}.
\newblock {\em Artificial Intelligence} 170:507--541.

\bibitem[\protect\citeauthoryear{Honda and
  Takemura}{2014}]{honda2014optimality}
Honda, J., and Takemura, A.
\newblock 2014.
\newblock {Optimality} of {Thompson} {Sampling} for {Gaussian} {Bandits}
  depends on {Priors}.
\newblock In {\em Artificial Intelligence and Statistics},  375--383.

\bibitem[\protect\citeauthoryear{Kaelbling, Littman, and
  Cassandra}{1998}]{kaelbling1998planning}
Kaelbling, L.~P.; Littman, M.~L.; and Cassandra, A.~R.
\newblock 1998.
\newblock {Planning} and {Acting} in {Partially} {Observable} {Stochastic}
  {Domains}.
\newblock {\em Artificial intelligence} 101(1):99--134.

\bibitem[\protect\citeauthoryear{Kaufmann, Korda, and
  Munos}{2012}]{kaufmann2012thompson}
Kaufmann, E.; Korda, N.; and Munos, R.
\newblock 2012.
\newblock {Thompson} {Sampling}: {An} {Asymptotically} {Optimal}
  {Finite}-{Time} {Analysis}.
\newblock In {\em International Conference on Algorithmic Learning Theory},
  199--213.
\newblock Springer.

\bibitem[\protect\citeauthoryear{Kocsis and
  Szepesv{\'a}ri}{2006}]{kocsis2006bandit}
Kocsis, L., and Szepesv{\'a}ri, C.
\newblock 2006.
\newblock {Bandit} based {Monte}-{Carlo} {Planning}.
\newblock In {\em ECML}, volume~6,  282--293.
\newblock Springer.

\bibitem[\protect\citeauthoryear{Lecarpentier \bgroup et al\mbox.\egroup
  }{2018}]{lecarpentier2018open}
Lecarpentier, E.; Infantes, G.; Lesire, C.; and Rachelson, E.
\newblock 2018.
\newblock {Open} {Loop} {Execution} of {Tree}-{Search} {Algorithms}.
\newblock In {\em Proceedings of the 27th International Joint Conference on
  Artificial Intelligence},  2362--2368.
\newblock IJCAI Organization.

\bibitem[\protect\citeauthoryear{Palacios and
  Geffner}{2009}]{palacios2009compiling}
Palacios, H., and Geffner, H.
\newblock 2009.
\newblock {Compiling} {Uncertainty} away in {Conformant} {Planning} {Problems}
  with {Bounded} {Width}.
\newblock {\em Journal of Artificial Intelligence Research} 35:623--675.

\bibitem[\protect\citeauthoryear{Perez~Liebana \bgroup et al\mbox.\egroup
  }{2015}]{perez2015open}
Perez~Liebana, D.; Dieskau, J.; Hunermund, M.; Mostaghim, S.; and Lucas, S.
\newblock 2015.
\newblock {Open} {Loop} {Search} for {General} {Video} {Game} {Playing}.
\newblock In {\em Proceedings of the 2015 Annual Conference on Genetic and
  Evolutionary Computation},  337--344.
\newblock ACM.

\bibitem[\protect\citeauthoryear{Phan \bgroup et al\mbox.\egroup
  }{2018}]{phan2018evade}
Phan, T.; Belzner, L.; Gabor, T.; and Schmid, K.
\newblock 2018.
\newblock {Leveraging} {Statistical} {Multi}-{Agent} {Online} {Planning} with
  {Emergent} {Value} {Function} {Approximation}.
\newblock In {\em Proceedings of the 17th International Conference on
  Autonomous Agents and Multiagent Systems}, AAMAS '18,  730--738.
\newblock Richland, SC: International Foundation for Autonomous Agents and
  Multiagent Systems.

\bibitem[\protect\citeauthoryear{Pineau, Gordon, and
  Thrun}{2006}]{pineau2006anytime}
Pineau, J.; Gordon, G.; and Thrun, S.
\newblock 2006.
\newblock {Anytime} {Point-based} {Approximations} for {Large} {POMDPs}.
\newblock {\em Journal of Artificial Intelligence Research} 27:335--380.

\bibitem[\protect\citeauthoryear{Powley, Cowling, and
  Whitehouse}{2017}]{powley2017memory}
Powley, E.; Cowling, P.; and Whitehouse, D.
\newblock 2017.
\newblock {Memory} {Bounded} {Monte} {Carlo} {Tree} {Search}.
\newblock {\em AAAI Conference on Artificial Intelligence and Interactive
  Digital Entertainment}.

\bibitem[\protect\citeauthoryear{Ross \bgroup et al\mbox.\egroup
  }{2008}]{ross2008online}
Ross, S.; Pineau, J.; Paquet, S.; and Chaib-Draa, B.
\newblock 2008.
\newblock {Online} {Planning} {Algorithms} for {POMDPs}.
\newblock {\em Journal of Artificial Intelligence Research} 32:663--704.

\bibitem[\protect\citeauthoryear{Silver and Veness}{2010}]{silver2010monte}
Silver, D., and Veness, J.
\newblock 2010.
\newblock {Monte}-{Carlo} {Planning} in {Large} {POMDPs}.
\newblock In {\em Advances in neural information processing systems},
  2164--2172.

\bibitem[\protect\citeauthoryear{Silver \bgroup et al\mbox.\egroup
  }{2016}]{silver2016mastering}
Silver, D.; Huang, A.; Maddison, C.~J.; Guez, A.; Sifre, L.; Van Den~Driessche,
  G.; Schrittwieser, J.; Antonoglou, I.; Panneershelvam, V.; Lanctot, M.;
  et~al.
\newblock 2016.
\newblock {Mastering} the {Game} of {Go} with {Deep} {Neural} {Networks} and
  {Tree} {Search}.
\newblock {\em Nature} 529(7587):484--489.

\bibitem[\protect\citeauthoryear{Silver \bgroup et al\mbox.\egroup
  }{2017}]{silver2017mastering}
Silver, D.; Schrittwieser, J.; Simonyan, K.; Antonoglou, I.; Huang, A.; Guez,
  A.; Hubert, T.; Baker, L.; Lai, M.; Bolton, A.; et~al.
\newblock 2017.
\newblock {Mastering} the {Game} of {Go} without {Human} {Knowledge}.
\newblock {\em Nature} 550(7676):354--359.

\bibitem[\protect\citeauthoryear{Smith and Simmons}{2004}]{smith2004heuristic}
Smith, T., and Simmons, R.
\newblock 2004.
\newblock {Heuristic} {Search} {Value} {Iteration} for {POMDPs}.
\newblock In {\em Proceedings of the 20th conference on Uncertainty in
  artificial intelligence},  520--527.
\newblock AUAI Press.

\bibitem[\protect\citeauthoryear{Somani \bgroup et al\mbox.\egroup
  }{2013}]{somani2013despot}
Somani, A.; Ye, N.; Hsu, D.; and Lee, W.~S.
\newblock 2013.
\newblock {DESPOT}: {Online} {POMDP} {Planning} with {Regularization}.
\newblock In {\em Advances in neural information processing systems},
  1772--1780.

\bibitem[\protect\citeauthoryear{Thompson}{1933}]{thompson1933likelihood}
Thompson, W.~R.
\newblock 1933.
\newblock {On} the {Likelihood} that {One} {Unknown} {Probability} exceeds
  {Another} in {View} of the {Evidence} of {Two} {Samples}.
\newblock {\em Biometrika} 25(3/4):285--294.

\bibitem[\protect\citeauthoryear{Weinstein and
  Littman}{2013}]{weinstein2013open}
Weinstein, A., and Littman, M.~L.
\newblock 2013.
\newblock {Open-loop} {Planning} in {Large}-{Scale} {Stochastic} {Domains}.
\newblock In {\em Proceedings of the Twenty-Seventh AAAI Conference on
  Artificial Intelligence},  1436--1442.
\newblock AAAI Press.

\bibitem[\protect\citeauthoryear{Yu \bgroup et al\mbox.\egroup
  }{2005}]{yu2005open}
Yu, C.; Chuang, J.; Gerkey, B.; Gordon, G.; and Ng, A.
\newblock 2005.
\newblock {Open}-{Loop} {Plans} in {Multi}-{Robot} {POMDPs}.
\newblock Technical report, Stanford CS Dept.

\end{thebibliography}

\end{document}